\documentclass[10pt,twocolumn,letterpaper]{article}

\usepackage{iccv}
\usepackage{times}
\usepackage{epsfig}
\usepackage{graphicx}
\usepackage{amsmath}
\usepackage{amssymb}
 \usepackage{longtable}
\usepackage{arydshln}
\usepackage{booktabs}
\usepackage{multirow}
\usepackage{multirow}
\usepackage{subfig}
\usepackage[pagebackref=true,breaklinks=true,letterpaper=true,colorlinks,bookmarks=false]{hyperref}
\usepackage{listings}
\usepackage{color}
\usepackage[ruled,vlined,linesnumbered]{algorithm2e}

\definecolor{codegreen}{rgb}{0,0.6,0}
\definecolor{codegray}{rgb}{0.5,0.5,0.5}
\definecolor{codepurple}{rgb}{0.58,0,0.82}
\definecolor{backcolour}{rgb}{1.0,1.0,1.0}

\iccvfinalcopy % *** Uncomment this line for the final submission

\def\iccvPaperID{4782} % *** Enter the ICCV Paper ID here

\ificcvfinal\fi
\newcommand{\modelname}{BUS }
\newcommand{\PretrainTaskName}{Patch-Text Matching }
\begin{document}

%%%%%%%%% TITLE
\title{\modelname: Efficient and Effective Vision-language Pre-training with Bottom-Up Patch Summarization. }

\author{Chaoya Jiang{$^1$}, Haiyang Xu{$^2\footnotemark[1]$}, Wei Ye{$^1\thanks{~~Corresponding Author.}$}, Qinghao Ye$^{2}$, Chenliang Li$^2$, Ming Yan$^2$, Bin Bi$^2$\\
 Shikun Zhang$^1$, Fei Huang$^2$, Songfang Huang$^2$\\
 $^1$National Engineering Research Center for Software Engineering, Peking University \\
$^2$DAMO Academy, Alibaba Group \\
% $^\heartsuit$ National University of Singapore\\
  {shuofeng.xhy@alibaba-inc.com, wye@pku.edu.cn}
  % \\
  % \texttt{\{chuanqi.tcq,f.huang\}@alibaba-inc.com}
}
\maketitle

%%%%%%%%% ABSTRACT
\begin{abstract}

% Large-scale pre-training of vision-language models has shown significant success in cross-modal tasks by learning cross-modal representations from image-text pairs. However, challenges persist due to the lengthy visual token sequences in ViT-based models, leading to ineffective and inefficient performance. In this paper, we propose a two-step approach inspired by document summarization in NLP to create a concise visual summary of a long visual token sequence guided by textual semantics. Our approach includes a Semantic-aware Patch Selector module (KPE) to perform coarse-grained Extractive visual summarization in the ViT backbone and a fine-grained generative summarization to obtain a further condensed visual representation sequence. We evaluate our approach on various VL understanding and generation tasks and show competitive or better downstream task performance while increasing efficiency by 40\%. Additionally, our model achieves well-designed SOTA downstream task performance by increasing input image resolution without increasing computational costs compared to baselines.

Vision Transformer (ViT) based Vision-Language Pre-training (VLP) models have demonstrated impressive performance in various tasks. However, the lengthy visual token sequences fed into ViT can lead to training inefficiency and ineffectiveness. Existing efforts address the challenge by either bottom-level patch extraction in the ViT backbone or top-level patch abstraction outside, not balancing training efficiency and effectiveness well. Inspired by text summarization in natural language processing, we propose a \textbf{B}ottom-\textbf{U}p Patch \textbf{S}ummarization approach named \modelname, coordinating bottom-level extraction and top-level abstraction to learn a concise summary of lengthy visual token sequences efficiently. Specifically, We incorporate a Text-Semantics-Aware Patch Selector (TSPS) into the ViT backbone to perform a coarse-grained visual token extraction and then attach a flexible Transformer-based Patch Abstraction Decoder (PAD) upon the backbone for top-level visual abstraction. This bottom-up collaboration enables our \modelname to yield high training efficiency while maintaining or even improving effectiveness. We evaluate our approach on various visual-language understanding and generation tasks and show competitive downstream task performance while boosting the training efficiency by 50\%. Additionally, our model achieves state-of-the-art performance on many downstream tasks by increasing input image resolution without increasing computational costs over baselines.
\end{abstract}
\vspace{-2ex}
\section{Introduction}
Large-scale pre-training of vision-language models has recently received tremendous success on a wide range of cross-modal tasks \cite{Tan2019LXMERTLC,Chen2020UNITERUI,Huang2020PixelBERTAI,Li2020OscarOA,Yu2021ERNIEViLKE,Li2021AlignBF,Wang2021SimVLMSV}. Such vision-language models learn cross-modal representations from a large number of image-text pairs by aligning the visual and linguistic modalities. 

Most recent works~\cite{Huang2020PixelBERTAI,Wang2021SimVLMSV,Li2021AlignBF,Kim2021ViLTVT,dou2021empirical} adopt ViT as the visual encoder or cross-modal fusion encoder, due to its excellent ability to model the fine-grained long visual sequences from the image grids or patches.

\begin{figure}[t]
     \centering
     \includegraphics[width=0.45\textwidth]{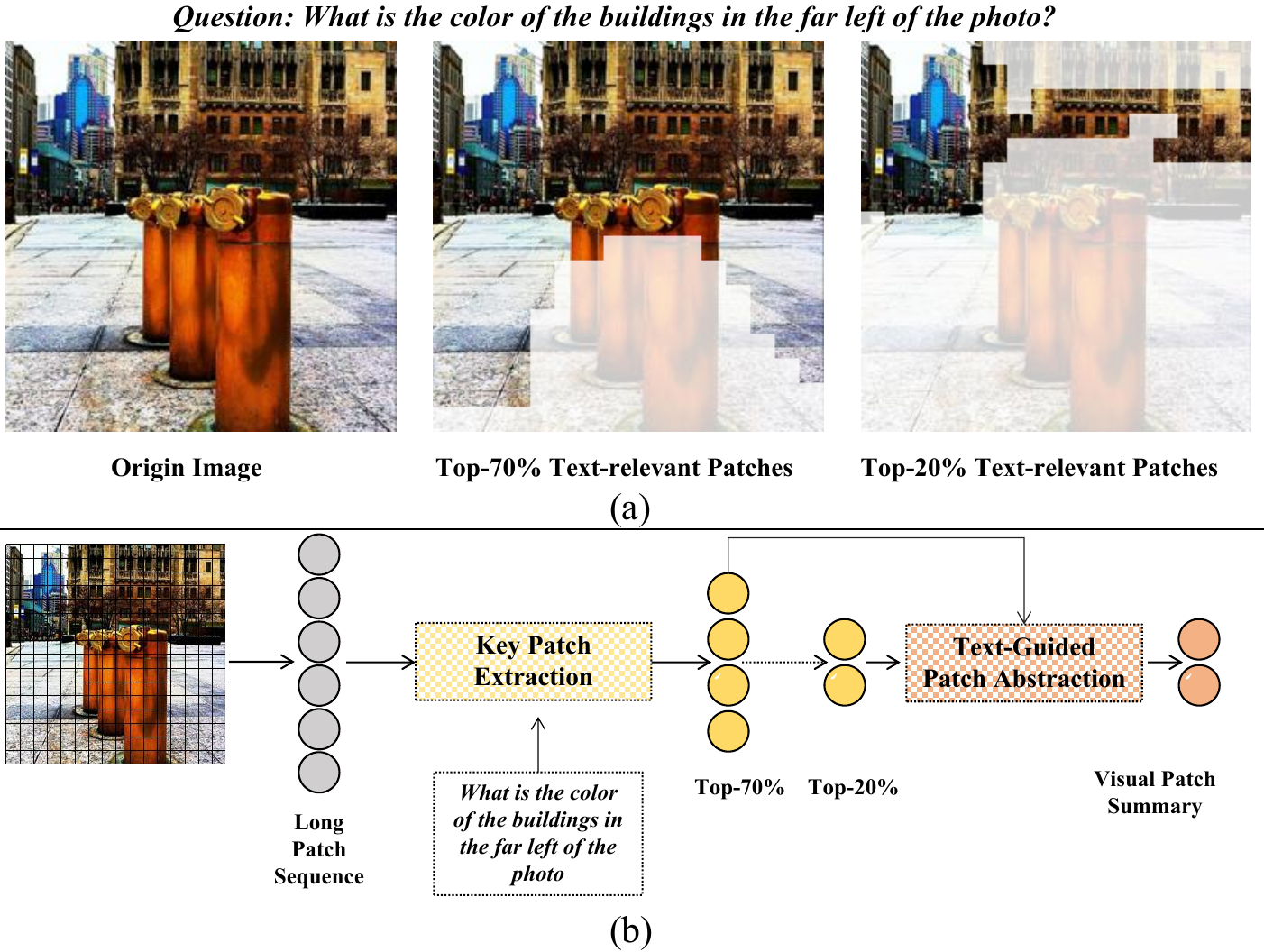}
   \vspace{-2ex}
     \caption{Subfigure (a)  shows the examples of selected text-relevant patch in the VQA scenario. Subfigure (b) shows the overview of our proposed bottom-up patch summarization.}
     \label{fig:fig1}
\vspace{-1ex}
\end{figure}

Despite the impressive progress of ViT-based VLP models, they still face challenges of training inefficiency and ineffectiveness caused by lengthy visual token sequences. Firstly, long visual sequences will bring heavy self-attention calculation for visual representation modeling and cross-modal fusion, leading to time-consuming training. Secondly, long visual sequences contain many redundant patches irrelevant to the text semantics. For instance, as illustrated in Figure \ref{fig:fig1}~(a), during the VQA task, when answering the question “What is the color of the buildings in the far left of the photo?", about 80\% of the image patches may be irrelevant with the question. On the one hand, those text-irrelevant patches (e.g., the yellow building in the image) will hinder the fine-grained alignment between the textual and visual modalities. On the other hand, they will lead to the overshadowing of brief linguistic signals (e.g., of short image captions) by complex visual ones during the cross-modal fusion, namely the “vanishing information" problem of textual information~\cite{li2022mplug}.

The limitations above underscore the importance of reducing visual token sequences. Recent related efforts can be categorized into two lines.

\begin{itemize}
     \item \textbf{Top-level Abstraction}. The first line tackles the issue outside the ViT-based visual backbones from a top-level perspective~\cite{Li2023BLIP2BL,Alayrac2022FlamingoAV}. Specifically, these works use a fixed number of learnable latent query vectors to query the long visual sequences output, obtaining the final fixed-length visual sequence representations, in an abstractive way. An obvious bottleneck is that they can not optimize the costly and potentially unnecessary self-attention calculation in the visual backbones. Meanwhile, this visual representation abstraction process only considers the semantics of the visual modality, ignoring the textual guidance and consequently leading to representation deficiency.

    \item  \textbf{Bottom-level Extraction}. The second line focuses on reducing patch tokens in the ViT-backbone from the bottom-level perspective, usually in an extractive manner~\cite{Rao2021DynamicViTEV,Liang2022EViTEV,Jiang2022TRIPSEV} . The problem here lies in that overly reducing the visual sequence by extracting critical tokens in the backbone, while accelerating the attention calculation, may deconstruct images' structural information. Therefore, balancing efficiency and effectiveness remains a bottleneck.
\end{itemize}

To achieve a better trade-off between the efficiency and effectiveness of VLP, we propose integrating the merits of top-level abstraction and bottom-level extraction. Inspired by bottom-up text summarization\cite{anderson1988teaching,jing1999decomposition}, which first select key phrases and then abstractively generate the final text summaries, we design a bottom-up summarization process for visual tokens. We first exploit coarse-grained key patch extraction in the ViT backbone, with regulation from text modality, to identify text-relevant tokens and remove potentially redundant ones, reducing the computational cost in the ViT backbone. Then fine-grained text-guided patch abstraction is performed upon the output sequence of the ViT backbone to obtain a further condensed visual representation sequence. 

Specifically, we incorporate a Text Semantic-aware Patch Selector (TSPS) module into the ViT-based backbone for bottom-level extraction. We transform object/region annotations to patch-level annotations to train an effective extractor with a novel auxiliary pre-training task named \PretrainTaskName (PTM), which facilitates patch extraction and fine-grained patch-text alignment. Next, we introduce a lightweight Transformer-based Patch Abstraction Decoder (PAD) for top-level abstraction. It takes the top-K text-relevant patch tokens from the output sequence of ViT-backbone as the input and the overall visual sequence as the encoder hidden states to generate the final visual patch summary.

We evaluate \modelname on various representative VL understanding and generation tasks, including visual question answering, cross-modal retrieval, and image captioning. We find that by reducing the length of the patch sequence to 20\% of its original length, we can not only get competitive or better downstream task performance but also enjoy a significant increase in efficiency over previous similar VLP models. For instance, \modelname reduces about 51\% of the inference time (see Table~\ref{table6}) and even improves by about 0.3 on the VQA test-dev with the same experimental settings. Furthermore, by increasing the input image resolution, \modelname achieves state-of-the-art downstream task performance (e.g., 78.28 on VQA test-dev) which benefits from processing more image tokens without increasing computational costs.

% Our contributions can be summarized as three-fold:
%  \begin{itemize}

% \item We propose an efficient vision-and-language pre-training model with Text-Relevant Image Patch Selection (TRIPS). As far as we know, this is the first exploration that decreases the computational cost of VLP models by reducing image tokens with the help of linguistic context.
% \item We propose a text-relevant patch-selection layer, which can dynamically compute text-dependent visual attention to identify the attentive image tokens and fuse inattentive ones with text guidance in an end-to-end manner.
% \item Extensive experiments indicate that our TRIPS can boost the VLP model efficiency at a lower computational cost than the un-accelerated baseline model. Furthermore, by increasing the input image resolution, TRIPS benefits from taking more image tokens to achieve better performance without increasing computational costs.
% \end{itemize}
\begin{figure*}
     \centering
     \includegraphics[width=0.97\textwidth]{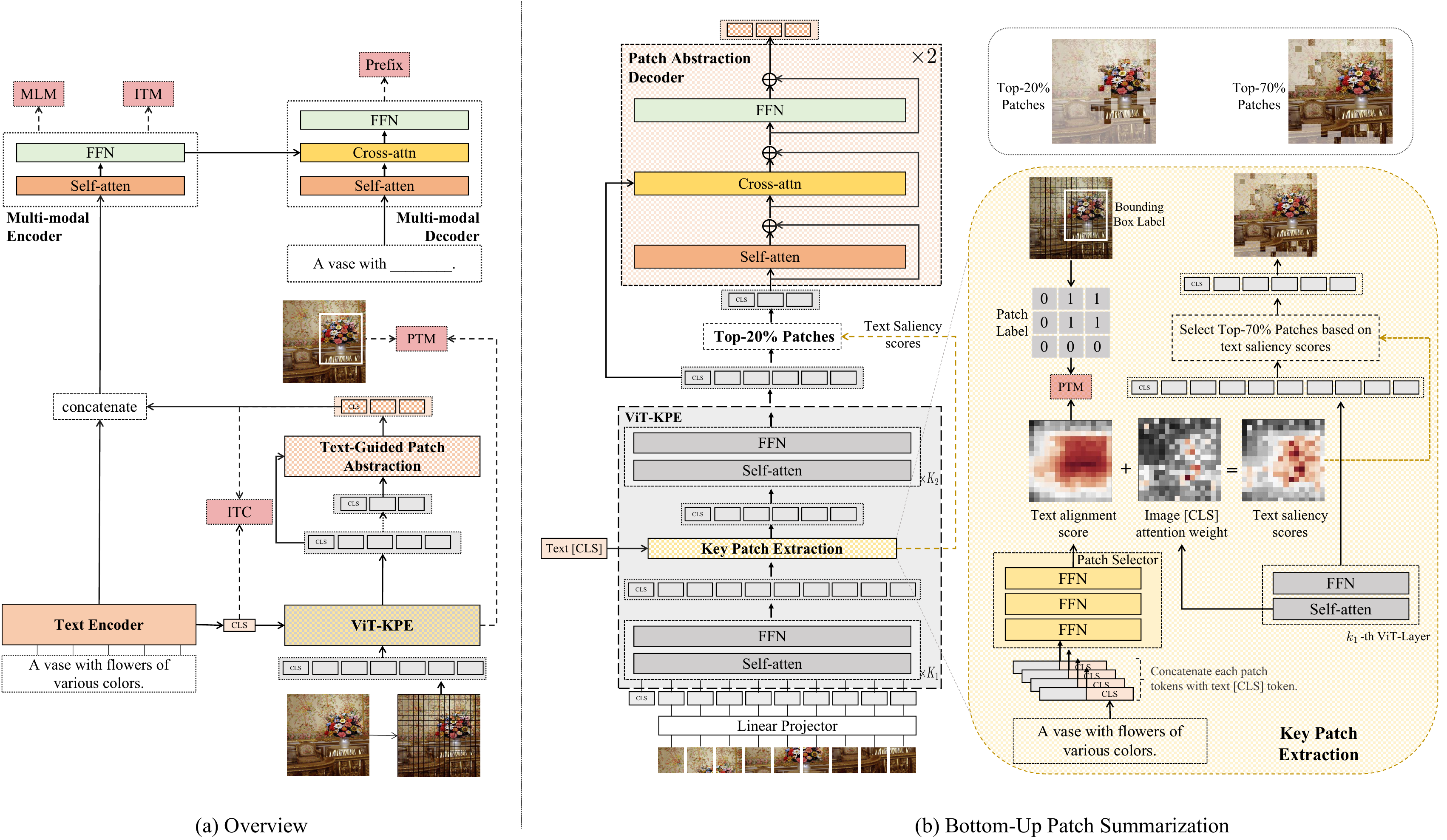}
     \caption{(a) Overview of our VLP model \modelname. (b) The details of the Bottom-Up Patch Summarization.}
     \label{fig:framework}
 \vspace{-1ex}
\end{figure*}
\section{Related Work}
\subsection{Vision-Language Pre-training}
% The existing work on vision language pre-training typically falls into two categories: Detector-based VLP model and CNN/ViT-Based VLP models.
% Previous Detector-based VLP methods \cite{Lu2019ViLBERTPT,Li2019VisualBERTAS, Tan2019LXMERTLC, Li2020OscarOA,Chen2020UNITERUI,Yu2021ERNIEViLKE} mainly take a two-step training pipeline approach, which first extracts visual features by a pre-trained object detector and then trains the cross-modal pre-training model to align text and visual features. Even though there are some region-based methods that reduce the computation cost with the lightweight model architecture \cite{Wang2020MiniVLMAS}, those methods still suffer from the expensive computational cost and time consumption for detecting objects/regions.  Recently, Vision Transformer (ViT) based \cite{Li2021AlignBF,Kim2021ViLTVT,Radford2021LearningTV,Wang2021VLMoUV,li2022blip,li2022mplug, Wang2021SimVLMSV,Kim2021ViLTVT} methods (especially the patch-based ViT) removes the complicated object detector in feature extraction to conduct end-to-end VL learning. These methods avoid the drawbacks of object detectors but face excessively long visual sequences without fine-grained cross-modal alignment information. Such long visual sequences also bring expensive computation costs and additional noise visual information for cross-modal fusion. In this paper, we propose a Bottom-Pp Patch Summarization approach coordinating bottom-level extraction and top-level abstraction to learn a concise summary of lengthy visual token sequences efficiently.
  Previous research on vision language pre-training has primarily fallen into two categories: Detector-based VLP models and CNN/ViT-based VLP models. Detector-based methods, such as  \cite{Lu2019ViLBERTPT,Li2019VisualBERTAS, Tan2019LXMERTLC, Li2020OscarOA,Chen2020UNITERUI,Yu2021ERNIEViLKE, Xu2023mPLUG2AM, Ye2023mPLUGOwlME}, use a two-step training pipeline that first extracts visual features using a pre-trained object detector and then aligns text and visual features using a cross-modal pre-training model. While some region-based methods, such as  \cite{Wang2020MiniVLMAS}, use lightweight model architectures to reduce computation costs, they still face expensive computational and time-consuming object detection. Recently, Vision Transformer (ViT) based methods, such as \cite{Li2021AlignBF,Kim2021ViLTVT,Radford2021LearningTV,Wang2021VLMoUV,li2022blip,li2022mplug, Wang2021SimVLMSV,Kim2021ViLTVT,Jiang2023COPAE,Jiang2023VisionLP,Jiang2022TRIPSEV} have emerged as a promising alternative to detector-based approaches. ViT-based models eliminate the need for object detectors in feature extraction, enabling end-to-end vision language learning. However, they struggle with lengthy visual token sequences and lack fine-grained cross-modal alignment information. These long visual sequences also increase computation costs and introduce noise visual information for cross-modal fusion. To address these challenges, we propose a Bottom-Up Patch Summarization approach that coordinates bottom-level extraction and top-level abstraction to efficiently learn a concise summary of lengthy visual token sequences.
 \subsection{Text Summarization}
   Existing text summarization methods can be roughly categorized into extractive summarization \cite{narayan2018ranking,xiao2019extractive,zhong2020extractive} and abstractive summarization \cite{Syed2021ASO,see2017get,paulus2018deep,zhang2020pegasus,Lewis2020BARTDS,Tan2017AbstractiveDS,Jiang2023ExploitingPI}. 
Extractive summarization extracts key sentences from the source document to form a summary. It can be formulated as a sentence classification problem \cite{narayan2018ranking,nallapati2017summarunner,Cheng2016NeuralSB}.
Abstractive summarization paraphrases important content after understanding the original document and constructs an abstract with newly generated words and coherent expressions.
Our proposed Bottom-Up patch Summarization is unpaired from the Text Summarization Task which contains a key patch extraction module and patch abstraction module.
\section{Method}

  In this section, we first provide an overview of our \modelname architecture. Then, we introduce the proposed Bottom-Up patch summarization which includes the Key Patch Extraction (KPE) incorporated in the Vision Transformer (ViT) backbone with a Text Semantic-aware Patch Selector (TSPS) and the Text-Guided Patch Abstraction (TPA) conducted on the output sequence of the ViT backbone. We leave the introduction of the pre-training task which includes the pre-training task \PretrainTaskName used to learn TSPS and the pre-training schedule in Appendix~\ref{sup:pretraining schedule}. 

\subsection{Model Architecture}
  As depicted in Figure~\ref{fig:framework}~(b), \modelname comprises a ViT-based image encoder with a Text Semantic-aware Patch Selector (TSPS) for coarse-grained Key Patch Extraction (KPE), a text encoder, a lightly Patch Abstraction Decoder(PAD) for fine-grained Text-Guided Patch Abstraction (TPA), a multi-modal fusion encoder for performing cross-modal interaction, and a multi-modal decoder for text generation. (Note that the text encoder contains 10 Transformer layers and PAD contains only 2 Transformer layers, which are initialized with BERT$_{base}$ \cite{Devlin2019BERTPO}).

Formally, let us consider an input image-text pair denoted as $\left(I, T\right)$. For the input text, we feed it to the text encoder and obtain the text representation $T = \{t_{cls}, t_1, t_2, \cdots, t_m \}$, where $t_{cls}$ is the embedding of the text [CLS] token used to summarize the global semantic information of the text. For the input image, we divide it into $n$ non-overlapping patches $P =\{p_{cls}, p_1,p_2, \cdots, p_{n} \}$. Then, we feed the patch sequence to the visual encoder. In the ViT backbone, we apply KPE to select text-relevant image patches that can reduce the visual sequence length and improve the training and inference efficiency of the ViT backbone. Suppose the output patch sequence representation of the ViT-backbone can be denoted as $V=\{v_{cls}, v_1, v_2, \cdots,v_u\}$, where $u\textless n$. Then, we feed the output patch sequence to PAD and conduct TPA to obtain the final visual summary of the long patch sequence, which can be denoted as $\hat{V}=\{\hat{v}_{cls}, \hat{v}_1, \hat{v}_2, \cdots,\hat{v}_s\}$, where $s \ll u$. The image and text representations are concatenated and fed into the cross-modal encoder. We obtain the cross-modal representations $\{c_{cls}, c_1, c_2, \dots, c_{l}\}$, where $l=s+m$. The cross-modal representations can be used to fine-tune downstream multi-modal understanding tasks. Additionally, the output cross-modal representations $\{c_{cls}, c_1, c_2, \dots, c_{l}\}$ of the multi-modal encoder are fed into a Transformer decoder for sequence-to-sequence learning
\subsection{Bottom-Up Patch Summarization}
  The proposed Bottom-Up Patch Summarization consists of two steps: Key Patch Extraction (KPE) and Text-Guided Patch Abstraction (TPA). KPE is conducted within the visual backbone to select a coarse-grained subset of text-relevant patch tokens. Then, outside the backbone, Text-Guided Patch Abstraction is conducted to abstract the selected tokens and obtain a more prominent visual summary. 
\vspace{-3ex}
\subsubsection{Key Patch Extraction}
\label{subsection:Key Patch Extraction }
\vspace{-0.5ex}
 Figure \ref{fig:framework}~(b) shows how we perform Key Patch Extraction (KPE) in the ViT-based visual backbone based on the Text Semantic-aware Patch Selector (TSPS). Similar to the text extractive summarization methods\cite{xiao2019extractive,zhong2020extractive}, we view the KPE as a patch classification task that aims to classify whether a patch is aligned with the text semantic and should be selected. Specially, suppose the TSPS is plugged between the $k_{th}$ ( $1\leq k  \textless N$ ) Transformer layer and $(k+1)_{th}$ Transformer layer and the output patch sequence features of $k_{th}$ Transformer layer is $v^{k}=\{v^{k}_{cls}, v^{k}_1, \cdots, v^{k}_n\}$. The text [CLS] feature is output by the text encoder and represents global information of the input text $T$. We will first concatenate the text [CLS] feature with each image patch token as follows:
$$
    \dot{v}^{k}_i = concat(v^{k}_i, t_{cls})
$$
where $v^{k}_i\in R^d, t_{cls}\in R^d, \dot{v}^{k}_i\in R^{2d}, i \in \{1, 2, \dots, n \}$.  Then the concatenated patch features ${\dot{v}^{k}_i}$ are fed to TSPS, which is a Multi-layer Perceptron (MLP) that contains three linear layers and is used to predict the alignment score between patches and the input text $T$.  The first two linear layers will linearly project the concatenated patch features $\{\dot{v}^{k}_i\}$ to the hidden representations $\{h^{k}_i\}$ and then the hidden representations $\{h^{k}_i\}$ is fed to the last linear layer denoted as $\mathbf{F}_{\theta}$ which can be seen as a classifier to predict whether the patches are relevant to the input text. The output of the last linear layer has only one dimension and will be fed to a Sigmoid activation function. Formally, the alignment score $a_i$ between the $i_{th}$ image patch and input text T can be calculated as follow:
$$
    a_{i} = Sigmoid(\mathbf{F}_{\theta}(h^{k}_i )), i \in \{1,2,\dots, n\}
$$
% To learn such effective patch selector, we transform bounding box annotations to generate the training patch-level labels (as discussed in Section \ref{sbusbusection2}) while also introduces additional bias. This is evident in Figure \ref{fig:framework}(b), where TSPS tends to predict all patches within the bounding box with a high alignment score, even though some of these patches may be irrelevant. To mitigate the bias, we also incorporate the attention map of the image [CLS] token to all patch tokens to help select the text-relevant patches based on the observation \cite{Caron2021EmergingPI} that the image [CLS] token in ViTs pays more attention (i.e.,having a larger attention value) to class-specific tokens than to the tokens on the non-object regions. By weighting the alignment socre $a$ and the attention map $p$ of the image [CLS] token to other tokens, we can highlight the patches corresponding to the objects in the detection box rather than other patches. This approach helps to alleviate the bias introduced by the bounding box labels and can be fomulated as follow:
To learn an effective patch selector, we generate patch-level training labels by transforming the bounding box annotations (as described in Appendix \ref{sup:pretraining Task}). However, this approach introduces bias, as illustrated in Figure \ref{fig:framework}~(b), where TSPS predicts all patches within the bounding box with a high alignment score, even if some of them are irrelevant. To address this issue, we incorporate the attention map of the image [CLS] token into the selection process. Previous work has shown that in ViTs, the image [CLS] token pays more attention to class-specific tokens than to tokens in non-object regions \cite{Caron2021EmergingPI}. By weighting the alignment score $a$ and the attention map $p$ of the image [CLS] token to other tokens, we can highlight the patches corresponding to objects in the detection box and suppress other patches. This approach reduces the bias introduced by the bounding box labels and can be formulated as follows:
$$
    \dot{a}_{i} = \beta * \mathbf{F}_{N} \left(a_{i}\right) + \left(1-\beta\right) * \mathbf{F}_{N} \left(p_{i}\right) 
$$
% where $p_{i}$ is attention value of the image [CLS] token to i-th patch tokens which is calculated in the $k_{th}$ transformer layer, $\dot{a}_{i}$ is named as the text salient score of the i-th patch in the sequence, $\beta $ is a hyper-parameter and $\mathbf{F}_{N}$ is a normalization function to normalize $p_{i}$ and $a_{i}$. 
Here, $p_{i}$ is the attention value of the image [CLS] token to the $i$-th patch token, calculated in the $k_{th}$ Transformer layer. We define $\dot{a}_{i}$ as the text saliency score of the $i$-th patch in the sequence, $\beta$ as a hyper-parameter, and $\mathbf{F}_{N}$ as a normalization function for $p_{i}$ and $a_{i}$.
 % Then, We select the top-$u$ image patch tokens from the patch sequence $\{v^{k}_{cls}, v^{k}_1, \cdots, v^{k}_n\}$ based on their text salient scores $\{a_{1}, ..a_{n}\}$, where $u = n \times \alpha$, and $ \alpha$ is a hyper-parameter which is used to control the proportion of selected patches to total patches.  
Then, we select the top-$u$ image patch tokens from the patch sequence $\{v^{k}_{cls}, v^{k}_1, \cdots, v^{k}_n\}$ based on their text saliency scores $\{\dot{a}_{1}, \cdots,\dot{a}_{n}\}$, where $u = n \times \alpha$ and $\alpha$ is the selection ratio that controls the proportion of selected patches to total patches. The selected top-$u$  image patch tokens are kept and we reconstruct the $k_{th}$ visual sequence as $v^{k} = \{v^{k}_{cls},v^{k}_{1}, \cdots,v^{k}_{u},{v}^{k}_{u}\}$. Then the reconstructed visual sequence is fed to the next $(k+1)_{th}$ Transformer layer. 
\vspace{-2ex}
\subsubsection{Text-Guided Patch Abstraction}
\vspace{-0.5ex}
While the coarse-grained key patch extraction removes some redundant visual tokens, many text-irrelevant tokens remain in the ViT backbone. However, we need to be cautious not to remove too many visual tokens in the ViT backbone, as this can lead to the loss of important structural information and affect the distribution of hidden representations in the backbone. To address this issue, we propose the Text-guided Patch Abstraction (TPA) outside the ViT backbone, using a lightweight Patch Abstraction Decoder (PAD). As shown in Figure \ref{fig:framework}~(b), the PAD consists of two transformer modules, each of which includes a self-attention layer and a cross-attention layer. Similar to \cite{Alayrac2022FlamingoAV} and \cite{Li2023BLIP2BL}, the model structure of PAD employs a similar way, but with a key difference: we select top-$s$ image patch tokens $\{\overline{v}_{cls},\overline{v}_{1},\cdots,\overline{v}_{s}\}$ from the output patch sequence $\{v_{cls},v_{1},\cdots,v_{s}\}$ of ViT backbone based on the text saliency scores $\{\dot{a}_{1}, \cdots, \dot{a}_{n}\}$, and take them as the input to guide the PAD to condense visual information while they used the static learnable embeddings without taking any prior of text information. Noted that $s = \gamma * u$ and $\gamma$ is the selection ratio for TPA. These top-$s$ patch tokens are highly relevant to the text semantics and provide a strong prior to help the PAD learn a condensed visual summary and highlight text-relevant visual information. After we select the top-$s$ image patch tokens $\{\overline{v}_{cls},\overline{v}_{1},\cdots,\overline{v}_{s}\}$, we will feed them to PAD. In each transformer module of the PAD, $\{\overline{v}_{cls},\overline{v}_{1},\cdots,\overline{v}_{l}\}$ first undergoes the self-attention layer, followed by a cross-attention with the output sequence $\{v_{cls},v_{1},\cdots,v_{s}\}$. Finally, the output of the PAD denoted as $\hat{V}=\{\hat{v}_{cls}, \hat{v}_1, \hat{v}_2, \cdots,\hat{v}_s\}$ serves as the final visual summary.
 
\section{Experiments}
\subsection{Data \& Setup}
\label{data_setup}
% Following the previous work~\cite{Li2021AlignBF}, we use the same pre-training dataset with 4M images with texts, which includes two in-domain datasets (MS COCO ~\cite{Lin2014MicrosoftCC} and Visual Genome ~\cite{Krishna2016VisualGC}), and
% three web out-domain datasets (Conceptual Captions ~\cite{sharma2018conceptualCA}, SBU Captions ~\cite{Ordonez2011Im2TextDI}. See Appendix \ref{sec: dataset} for more details.

% We pre-train the model for 30 epochs with a total batch size of 1024 on 16 NVIDIA A100 GPUs. We use a 6-layer Transformer for both the text encoder and the cross-modal skip-connected network, and a 12-layer Transformer for the decoder. The text encoder is initialized using the first 6 layers of the BERT$_{base}$~\cite{Devlin2019BERTPO} model and the skip-connected network is initialized using the last 6 layers of the BERT$_{base}$.
  In our work, we follow the pre-training setup established in \cite{Li2021AlignBF} and utilize the same pre-training dataset consisting of 4 million images with associated texts. The dataset comprises two in-domain datasets, MS COCO~\cite{Lin2014MicrosoftCC} and Visual Genome~\cite{Krishna2016VisualGC}, as well as three out-domain datasets, Conceptual Captions~\cite{sharma2018conceptualCA}, and SBU Captions~\cite{Ordonez2011Im2TextDI}. Further details on the dataset can be found in Appendix \ref{sup:pretraining data}.

 We pre-train the model for 30 epochs with a total batch size of 1024 on 8 NVIDIA A100 GPUs. We use a 10-layer Transformer for the text encoder, a 2-layer transformer for the patch abstraction decoder,  a 3-layer for the cross-modal encoder network, and a 12-layer Transformer for the cross-modal decoder. Specifically, we initialize the text encoder using the first 10 layers of the BERT${base}$~\cite{Devlin2019BERTPO} model, initialize the patch abstraction decoder using the last 2 layers of BERT${base}$, initialize the cross-modal encoder network using the last 3 layers of BERT${base}$.
 
\begin{table*}[t]
\setlength\tabcolsep{6.5pt}
\centering
\footnotesize
\begin{tabular}{l|c|cc|cccccccc|cc}
\toprule[2.0pt]
\multicolumn{1}{c|}{\multirow{3}{*}{Models}}      &
%\multirow{2}{*}{\# Pre-train} &
\multicolumn{1}{c|}{\# Pre-train} &
\multicolumn{2}{c|}{VQA} &
\multicolumn{8}{c|}{COCO Caption} & \multicolumn{2}{c}{\multirow{1}{*}{NoCaps}}  \\
\multicolumn{1}{c|}{\multirow{2}{*}{}}      &
% \multirow{2}{*}{\# Pre-train} &
\multicolumn{1}{c|}{\multirow{2}{*}{Data}} & \multicolumn{2}{c|}{} &
\multicolumn{4}{c}{Cross-entropy Optimization} & \multicolumn{4}{c|}{CIDEr Optimization} & \multicolumn{2}{c}{}  \\
      &   & Test-std  & Test-dev & B@4 & M & C & S & B@4 & M & C & S & C & S     \\
      
\midrule      
E2E-VLP~\cite{Xu2021E2EVLPEV} & 4M & 73.25  & 73.67 & 36.2 &-&117.3&-&  - & - & - & - & - & - \\
OSCAR~\cite{Li2020OscarOA} & 6.5M & 73.16 & 73.44 & - & - & - & - & 41.7 & 30.6 & 140.0 & 24.5 & 83.4 & 11.4 \\
VinVL~\cite{2021VinVL} & 5.65M & 76.52 & 76.60 & 38.5 & 30.4 & 130.8 & 23.4 & 41.0 & 31.1 & 140.9 & 25.2 & 97.3 & 13.8 \\

% SimVLMhuge_{huge} \cite{wang2021simvlm} & 1.8B & 40.6 & 33.7 & 143.3 & 25.4 & - & - & - & - & 112.2 & -  \\
% LEMONlarge_{large} & 200M & - & - & 40.6 & 30.4 & 135.7 & 23.5 & 42.3 & 31.2 & 144.3 & 25.3 & 113.4 & \textbf{15.0} \\
METER~\cite{dou2021empirical} & 4M & 77.68 & 77.64  & - & - & - & - & - & - & - & - & - & - \\
BLIP~\cite{li2022blip}  & 14M & 77.54 & 77.62 & 38.6 & - & 129.7 & - & - & - & - & - & \textbf{105.1} & \textbf{14.4}  \\
% VLMo & - & 79.94 & 79.98  & - & - & - & - & - & - & - & - & - & - \\
SimVLM~\cite{Wang2021SimVLMSV} & 1.8B & 77.87 & 78.14  & 39.0 & \textbf{32.9} & \textbf{134.8} & 24.0 & - & - & - & - & - & - \\
ALBEF~\cite{Li2021AlignBF} & 4M & 74.54 & 74.70  & - & - & - & - & - & - & - & - & - & - \\
ALBEF~\cite{Li2021AlignBF} & 14M & 75.84 & 76.04  & - & - & - & - & - & - & - & - & - & - \\
% OFA  & 18M & 79.87 & 80.02 & - & - & - & - & 43.5 & 31.9 & 149.6 & \textbf{26.1} & - & - \\
% SimVLMlarge_{large} & 1.8B &  80.03 & 80.34 & 40.3 & \textbf{33.4} & \textbf{142.6} & \textbf{24.7} & - & - & - & - & - & - \\
% Florence & 0.9B & 80.16&80.36 & - & - & - & - & - & - & - & - & - & - \\
XVLM~\cite{Zeng2021xvlm} & 4M & 78.07 & 78.09 & 39.8 & - & 133.1 & -& 41.3 & - & 140.8 & -  & - & - \\
mPLUG~\cite{li2022mplug} & 4M & 77.55  & 77.73 & 39.3 & 30.1 & 132.4 & 23.34 & 41.2 & 30.8 & 140.2 & 25.2  & 98.3 & 12.9 \\
\midrule

\modelname$_{384}$ & 4M & 77.89  & 77.98 & 39.5 & 30.9 & 132.6 & 23.93 & 41.4 & 31.0 & 140.7 & 25.3  & 98.8 & 12.9  \\
\modelname$_{512}$ & 4M & \textbf{78.28}  & \textbf{78.34} & \textbf{40.04} & 31.3 & 133.6 & \textbf{24.12} & \textbf{41.8} & \textbf{31.4} & \textbf{141.1} & \textbf{25.9} & 99.1 & 13.2 \\
\bottomrule[2.0pt]
\end{tabular}
\vspace{-2ex}
%& PixelBERT-r50   & 71.35    & 71.42        & 59.80 & 85.50 & 91.60            & 75.70 & 94.70 & 97.10             \\
%& PixelBERT-x152  & 74.45    & 74.55        & 71.50 & 92.10 & 95.80            & 87.00 & \textbf{98.90} & \textbf{99.50}             \\
%& VILLA-base   & 73.59    & 73.67        & 74.74 & 92.86 & 95.82            & 75.70 & 94.70 & 97.10             \\
%& VILLA-large  & 74.69    & 74.87        & 76.26 & 94.24 & 96.84            & 87.90 & 97.50 & 98.80             \\
\caption{Evaluation Results on VQA, image captioning on COCO Karpathy test split \cite{karpathy2015deep} and NoCaps \cite{nocaps} validation set. B@4: BLEU@4, M: METEOR, C: CIDEr, S: SPICE. More details about comparison models in Appendix \ref{sup:comparison models}, \modelname$_{384}$ means we set image resolution to $384 \times 384$ during finetuning. Similar, \modelname$_{512}$ means we set image resolution to $512 \times 512$. } 
\label{table:vqa_caption}
\vspace{-1ex}
\end{table*}

\begin{table*}[t]
\setlength\tabcolsep{6.7pt}
\centering
\footnotesize
\begin{tabular}{l|c|cccccc|cccccc}
\toprule[2.0pt]
\multicolumn{1}{c|}{\multirow{2}{*}{Models}}      &
%\multirow{2}{*}{\# Pre-train} &
\multicolumn{1}{c|}{\# Pre-train} &
\multicolumn{6}{c|}{MSCOCO (5K test set)} & \multicolumn{6}{c}{Flickr30K (1K test set)} \\
      &  data & \multicolumn{3}{c}{TR} & \multicolumn{3}{c|}{IR} & \multicolumn{3}{c}{TR} & \multicolumn{3}{c}{IR}          \\
\midrule
&&R@1&R@5&R@10&R@1&R@5&R@10&R@1&R@5&R@10&R@1&R@5&R@10 \\ \hline
ALIGN~\cite{jia2021scaling} & 1.8B  & 77.0&93.5&96.9&59.9&83.3&89.8&95.3& 99.8&100.0&84.9&97.4&98.6   \\
OSCAR~\cite{Li2020OscarOA} & 4M  & 70.0&91.1&95.5&54.0&80.8&88.5&-& -&-&-&-&-   \\
E2E-VLP~\cite{Xu2021E2EVLPEV} & 4M     &-& -&-&-&-&- & 86.2 &97.5 &98.92&73.6 & 92.4 &96.0 \\
UNITER~\cite{Chen2020UNITERUI} & 4M     & 65.7&88.6&93.8&52.9&79.9&88.0&87.3& 98.0&99.2&75.6&94.1&96.8  \\
VLMo~\cite{Wang2021VLMoUV} & 4M & 78.2& 94.4& 97.4& 60.6& 84.4& 91.0& 95.3& 99.9& 100.0& 84.5& 97.3& 98.6 \\
ALBEF~\cite{Li2021AlignBF} & 14M & 77.6&94.3&97.2&60.7&84.3&90.5&95.9& 99.8&100.0&85.6&97.5& \textbf{98.9}                 \\
% Florence & 0.9B & 81.8&95.2&-&63.2&85.7&-&97.2& 99.9&-&87.9&98.1&-                 \\
% BLIP & 14M & 80.6 &95.2&97.6&63.1&85.3&91.1&96.6& 99.8&100.0&87.2&97.5&98.8                 \\
XVLM~\cite{Zeng2021xvlm} & 4M & 80.4 & 95.5&\textbf{98.2} & 63.1 & \textbf{85.7} &\textbf{ 91.6} & 96.8 & 99.8 & 100.0 & 86.1 & 97.4 &98.7\\
mPLUG~\cite{li2022mplug} & 4M  & 80.5 &95.4&97.9&63.3&85.3&91.2&96.7& 99.8&100.0& 86.5&97.5&98.8   \\
\midrule
% \modelname & 4M  & 80.6 &95.2&97.6&63.1&85.3&91.1&96.6& 99.8&100.0&87.2&97.5&98.8  
\modelname & 4M  & \textbf{80.6} & \textbf{95.7} & 98.0 & \textbf{63.6} &85.5& 91.5&\textbf{97.0}& 99.8&100.0&\textbf{86.9} & \textbf{97.8} & 98.8   \\
\bottomrule[2.0pt]
\end{tabular}      
\vspace{-2ex}
\caption{Evaluation results of image-text retrieval on Flickr30K~\cite{Plummer2015Flickr30kEC} and COCO datasets~\cite{Lin2014MicrosoftCC}.}
\label{table:retrieval}
\vspace{-4ex}
\end{table*}

\begin{table}[t]
\centering
\footnotesize
 \setlength{\tabcolsep}{5.3mm}{
\begin{tabular}{@{}lccccc@{}}
\toprule[2.0pt]
\multicolumn{1}{c}{\multirow{2}{*}{Model}} & \multicolumn{3}{c}{RefCOCO+} \\
\multicolumn{1}{c}{}          & testA   & testB   & val          \\ \midrule

UNITER~\cite{Chen2020UNITERUI}  & 75.90    & 81.45   & 66.70         \\
VL-BERT~\cite{Su2020VLBERTPO} &72.59  & 78.57 & 62.30   \\
ViLBERT~\cite{Lu2019ViLBERTPT} & 72.34&  78.52 &   62.61             \\
VILLA~\cite{gan2020large}   & 76.17    & 81.54   & 66.84            \\
MDETR~\cite{Kamath2021MDETRM}    & 79.52    & 84.09   & 70.62      \\
UNICORN~\cite{Yang2021CrossingTF}  & 80.30    & 85.05   & \textbf{71.88}      \\
XVLM~\cite{Zeng2021xvlm}     & 80.17  &  86.36  & 71.00 \\ 
mPLUG~\cite{li2022mplug}    & 80.07  &   85.21 &  71.03       \\ \hline
\modelname$_{384}$   & 80.11  &   86.03 &  71.21       \\ 
\modelname$_{512}$    & \textbf{80.36}  &  \textbf{ 86.61} &  71.84       \\ 
\bottomrule[2.0pt]
\end{tabular}}
\vspace{-1ex}
\caption{Evaluation results of visual grounding on ReferCOCO+. We use the accuracy of IOU 0.5 on visual grounding (a prediction is right if the IoU between the grounding-truth box and the predicted bounding box is larger than 0.5)}
\label{tab:visual_grounding}
 \vspace{-3ex}
\end{table}
\vspace{-0.5ex}
\subsection{Main Result}

% We evaluate our model \modelname on four widely explored vision-language downstream tasks: Visual Question Answering (VQA), Cross-modal Retrieval, Image Caption, and Visual Grounding (VG). For the proposed model\modelname, we conduct the Extractive patch summarizartion after the 6-th Transformer layer in the ViT encoder and select about 70\% text-relevant patch tokens, for the fine-grained patch summarization, we select Top-20\% text relevant patch tokens as the input for the lightly summarization decoder. As the experiments in the subsection \ref{subsection:Impact_of_setting} suggests that such setting achieve the desired trade-off between the downstream task performance and the model inference speed. The datasets and fine-tuning hyperparameters are detailed in Appendix \ref{sec:downsteam}. Details of the comparison methods are in Appendix \ref{sec:compare}.
  We evaluate the effectiveness of our proposed model, \modelname, on four well-established vision-language downstream tasks: Visual Question Answering (VQA), Cross-modal Retrieval, Image Captioning, and Visual Grounding (VG). The Key Patch Extraction (KPE) is performed after the 6th Transformer layer in the ViT encoder, where about 70\% text-relevant patch tokens are selected after the coarse-grained extraction. For the Text-Guided Patch Abstraction (TPA), we select the top 20\% of text-relevant patch tokens as the input for the Patch Abstraction Decoder (PAD). Our experiments in subsection \ref{subsection:Impact_of_setting} and subsection \ref{subsection:Impact_of_setting} demonstrate that this setting achieves the desired trade-off between downstream task performance and model inference speed. Please refer to Appendix \ref{sup:pretraining data}  for more information on the datasets and fine-tuning hyper-parameters, and to Appendix \ref{sup:comparison models} for details on the comparison methods used in the experiments.
\vspace{-1ex}
\subsubsection{Visual Question Answering}

The VQA task \cite{Agrawal2015VQAVQ} requires the model to answer natural language questions given an image.  Following the approach proposed in \cite{Li2021AlignBF}, we treat VQA as an answer-generation problem. We evaluate the performance of our proposed model, \modelname, by submitting our results to the evaluation server \footnote{https://eval.ai/web/challenges/challenge-page/830/overview} in Table \ref{table:vqa_caption}. and report the test-dev and test-std scores in Table \ref{table:vqa_caption}. Our results show that \modelname achieves comparable performance with state-of-the-art models under the same image resolution (384 $\times$ 384) while being about 50\% faster in terms of model inference time (as reported in subsection~\ref{HigherImages}). Furthermore, when we increase the image resolution to 512 $\times$ 512, our model achieves state-of-the-art performance while maintaining a similar inference computation cost to other baselines. These results demonstrate the effectiveness and efficiency of \modelname in VQA tasks.

% We follow \cite{Li2021AlignBF} and consider VQA as an answer-generation problem. We report test-dev and test-std scores by submitting our results to the evaluation 
% server\footnote{https://eval.ai/web/challenges/challenge-page/830/overview} in Table \ref{table:vqa_caption}. Compared with the VLP baselines, our \modelname can get the comparable performance with SOTAs under the same image resolution (384 $\times$ 384) but speed up about 40\% of model inference(see the report result in subsection~\ref{HigherImages}). When we increase the image resolution to 512 $\times$ 512, we can achieve SOTA performance while keeping a similar inference computation cost with other baselines. The results demonstrate the effectiveness and efficiency of \modelname.
\vspace{-2ex}
\subsubsection{Image Captioning} 
\vspace{-1ex}
% As there is no textual input in the image caption task, we directly set the hyperparameter $\beta$ to 0 and select the patches based on the attention weight of the image [CLS] token to other image tokens. Following~\cite{Li2020OscarOA}, we first fine-tune \modelname with cross-entropy loss and then with CIDEr optimization~\cite{scst} for extra 5 epochs. As shown in Table~\ref{table:vqa_caption}, when we set the image resolution to $384 \times 384$, \modelname can still get the comparable result with SOTA models, including XVLM and BILIP on both COCO Caption and Nocaps datasets. When we set  the image resolution to $512 \times 512$.\modelname performs the best on CIDEr evaluation and surpasses the SOTA model.
For the image captioning task, where there is no textual input, we set the hyper-parameter $\beta$ to 0 and select patches based on the attention weight of the image [CLS] token to other image tokens. Following~\cite{Li2020OscarOA}, we fine-tuned \modelname with cross-entropy loss and then with CIDEr optimization for an extra 5 epochs. Our experiments, as shown in Table~\ref{table:vqa_caption}, demonstrate that \modelname achieves comparable results with SOTA models when the image resolution is set to $384 \times 384$. Moreover, when we set the image resolution to $512 \times 512$, \modelname performs even better on CIDEr evaluation, surpassing state-of-the-art models.
% \begin{table}[htbp]
% \small
% \setlength{\tabcolsep}{1.6mm}
% \begin{tabular}{ccccc}
% \toprule[2.0pt]
% Models & Params (M) & VQA & FLOPs (G) & Throughput  \\ \hline
% ViLBERT  &274.3&   70.55 &958.10 &  6.40\\
% VisualBERT  &170.3& 70.80  &425.02 &  6.10\\
% LXMERT  &239.8&  72.42  &952.00  &  6.38\\
% UNITER  &154.7 & 72.70  &    949.90    &   6.42      \\
% OSCAR   &154.7  & 73.16   &    956.40  &    6.35          \\
% VinVL   &157.3 &  76.52   &    1023.30    &    7.32             \\
% E2E-VLP & 94 & 73.25  &    144.30     &     80.23        \\
% ViLT     & 87.4&  71.26 &    55.40   &  247.530           \\ 
% ALBEF    &210&  74.54  &33.42    &      197.52     \\
% XVLM   & 216& \textbf{78.07}   & 38.65    &     174.42      \\
% TRIPS   & 210 &   76.23 &  20.89 &   343.05        \\
% mPLUG   & 220 & 77.55  & 36.63    &      186.42      \\\hline
% \modelname  &210  & 77.86 &   \textbf{ 19.03 } &      \textbf{360.23}      \\  \bottomrule[2.0pt]
% \end{tabular}
% \centering
% \caption{The comparison of the efficiency of different models. Here, we report the VQA test-dev result and FLOPs, throughput and latency. The FLOPs results of the baselines come from  \cite{Kim2021ViLTVT}. Since FLOPs are proportional to input size, for a fair comparison, we keep same the input size with \cite{Kim2021ViLTVT}, which is 197 for image patches (the image resolution is $224 \times 224$ ) length and 40 for text tokens length. We keep the same setting when calculating throughput and latency.}
% \label{table_efficient}
% \end{table}

\vspace{-2ex}
\subsubsection{Image-Text Retrieval}
\vspace{-1ex}
% We conduct experiments for both image-to-text retrieval (TR) and text-to-image retrieval (IR) on MSCOCO \cite{Lin2014MicrosoftCC} and Flickr30K \cite{Plummer2015Flickr30kEC} datasets. We jointly optimize the ITC loss and the ITM loss during fine-tuning. 
% The results are reported in Table \ref{table:retrieval}. As shown in Table \ref{table:retrieval}, the experimental results show that our model gets comparable performance with other VLP baselines. 
We conduct experiments on MSCOCO \cite{Lin2014MicrosoftCC} and Flickr30K \cite{Plummer2015Flickr30kEC} datasets for both image-to-text retrieval (TR) and text-to-image retrieval (IR), and jointly optimize the ITC loss and the ITM loss during fine-tuning. The results are reported in Table \ref{table:retrieval}. Our model demonstrates comparable performance with other VLP baselines, as shown in the experimental results.
\vspace{-2ex}
\subsubsection{Visual Grounding} 
% Table \ref{tab:visual_grounding} shows that when we set the image resolution to $384 \times 384$, \modelname get the comparable result compared with SOTA methods.  Due to the elimination of the undetected redundant patches in the visual backbone of the model, we can improve the image resolution and get a better result without increasing more computational costs compared with other methods. When we increase the image resolution to $512 \times 512$, our model outperforms all the SOTA methods, which indicates the efficiency and effectiveness of \modelname.
Table \ref{tab:visual_grounding} demonstrates the performance of \modelname in the visual grounding task. When the image resolution is set to $384 \times 384$, our model achieves comparable results with competitive baseline methods. Thanks to our bottom-up summarization mechanism, we are able to improve the image resolution and obtain better results without incurring additional computational costs compared to other methods. When we increase the image resolution to $512 \times 512$, our model outperforms all state-of-the-art methods, demonstrating the effectiveness and efficiency of \modelname.
% \subsubsection{Natural Language for Visual Reasoning}

% The NLVR2 \cite{Suhr2019ACF} task requires the model to predict whether a sentence describes a pair of images which is a binary classification task. 
%  We follow \cite{Li2021AlignBF} and use two cross-attention layers to process the two input images, and their outputs are merged and fed to a Feed Forward Network (FFN). An MLP classifier is then applied to the output embedding of the text [CLS] token.  
% As shown in Table ?\ref{table1}, \modelname has a better performance than existing VLP methods.

\begin{figure}
     \centering
     \includegraphics[width=0.46\textwidth]{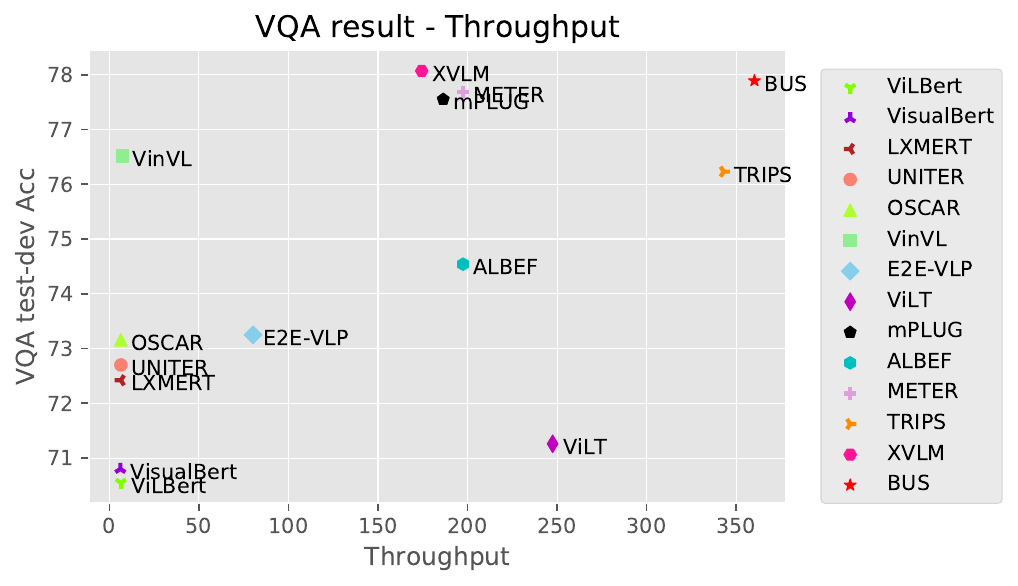}
     \caption{The visualization of the VQA test-dev result and Throughput of different VLP models.}
     \label{fig:efficiency}
\end{figure}

\subsection{Efficiency of \modelname}

% To investigate the efficiency of the proposed Bottom-Up Patch Summerization mechanism, we first compare the computational complexity of different models. We report the Floating Point Operations Per second (FLOPs), a widely used evaluation metric for model computational complexity. Besides, to evaluate the computational speed of our model, we compare the throughput and latency of different models. We use a Xeon Platinum 8163 CPU and an NVIDIA V100 GPU to calculate the latency and throughput. As shown in Table \ref{table_efficient}, \modelname not only has the lowest computational complexity (e.g., 24.04 of FLOPs) but also the fastest computational speed (e.g., 313.71 of Throughput and 14ms of Latency).
  To investigate the efficiency of the \modelname, we conduct the following experiments. We first compare the computational complexity of recent SOTA VLP models and report the Floating Point Operations Per second (FLOPs) which is a widely used evaluation metric for model computational complexity. In addition, we evaluate the computational speed of our model by comparing the throughput and latency of different models. We use a Xeon Platinum 8163 CPU and an NVIDIA V100 GPU to calculate the latency and throughput. As shown in Table \ref{table_efficient} and Figure \ref{fig:efficiency}, our \modelname not only has the lowest computational complexity (e.g., 19.03 of FLOPs) but also the fastest computational speed (e.g., 360.23 Throughput and 14ms Latency).  
\begin{table}[htbp]
\footnotesize
\setlength{\tabcolsep}{3mm}
\begin{tabular}{cccc}
\toprule[2.0pt]
Models & Latency (ms)  &  FLOPs (G) & Throughput   \\ \hline
% ViLBERT~\cite{Lu2019ViLBERTPT} &274M&   958.10 &  6.40 \\
% VisualBERT~\cite{Li2019VisualBERTAS}  &170M&425.02 &  6.10 \\
% LXMERT~\cite{Tan2019LXMERTLC}  &240M&  952.00  &  6.38  \\
UNITER \cite{Chen2020UNITERUI}  & 870ms &     949.90    &   6.42      \\
OSCAR \cite{Li2020OscarOA}  & 860ms &     956.40  &    6.35          \\
VinVL \cite{2021VinVL}  & 640ms &    1023.30    &    7.32             \\
E2E-VLP \cite{Xu2021E2EVLPEV} &  70ms&    144.30     &     80.23        \\
ViLT \cite{Kim2021ViLTVT}    &  19ms&     55.40   &  247.53          \\ 
ALBEF \cite{Li2021AlignBF}   & 22ms&  33.42    &      197.52     \\
XVLM  \cite{Zeng2021xvlm} &  27ms& 38.65    &     174.42      \\
TRIPS  \cite{Jiang2022TRIPSEV} & 11ms &  20.89 &   343.05        \\
mPLUG \cite{li2022mplug}  & 24ms&  36.63    &      186.42      \\\hline
\modelname  & \textbf{10}ms& \textbf{ 19.03 } &      \textbf{360.23}      \\  \bottomrule[2.0pt]
\end{tabular}
\centering
\vspace{-1ex}
\caption{The comparison of the efficiency of different models. FLOPs, throughput, and latency are reported here. Since FLOPs are proportional to input size, for a fair comparison, we use same the input size with \cite{Kim2021ViLTVT}, which is 197 for image patches length and 40 for text tokens length. We also keep the same setting when calculating throughput and latency.}
\label{table_efficient}
\vspace{-1ex}
\end{table}

\begin{table}[h]
\footnotesize
 \setlength{\tabcolsep}{2.0mm}{
\begin{tabular}{cccccccc}
\toprule[2.0pt]
 Locs & SR-KPE & SR-TPA     & VQA & FLOPs (G) & Throughput  \\ \hline
4   & 40\%  &  20\%      &    76.22   &   16.22  &  443.48 \\
4   &70\%   &  20\%      &    77.19   &   18.53  & 381.65  \\ 
4   &90\%   &  20\%      &    77.38   &  20.04   & 338.12  \\ \hline
6  & 40\%   &  20\%      &  76.87     &   17.24   & 410.05  \\
6  & 70\%   &  20\%     &  77.89      &  19.03   & 360.23\\
6  & 90\%   &  20\%     &  77.92      &  21.77    &  316.11 \\\hline
8 & 40\%    &  20\%      &  76.97     &  19.09   & 353.13  \\
8  & 70\%   &  20\%    &  77.94       &  21.65      &  312.06 \\
8 & 90\%    &  20\%    &   78.01      & 23.24      & 242.88 \\ \bottomrule[2.0pt]

\end{tabular}}
\vspace{-2ex}
\centering
\caption{Results of pre-training and fine-tuning \modelname with different selection locations and selection ratios. We report the text-dev score results of VQA, FLOPs, and Throughput. SR-KPE refers to the Selection Ratio for Key Patch Extraction, and SR-TPA refers to the Selection Ratio for Text-Guided Patch Abstraction.}
\label{table:impactofKR}
\vspace{-3ex}
\end{table}

% \subsection{Effectiveness of Bottom-Up Patch Summerization}

\subsection{ The Impact of Selection Ratio for Text-Guided Patch Abstraction }
\begin{figure}[!t]
% \subfigure[ R@1 of Text Retrieval]{
% \includegraphics[width=0.24\textwidth]{iccv2023AuthorKit/vqa_throughput_pa_locat (2).pdf}
% }%
\centering
\includegraphics[width=0.42\textwidth]{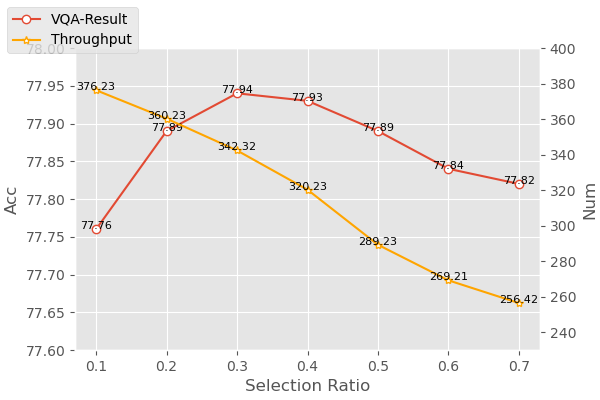}
\vspace{-2ex}
\caption{VQA performance and throughput of \modelname on different selection ratio of Text-Guided Patch Abstraction.}
\label{fig:impact of input number}
 \vspace{-3ex}
\end{figure}

  To investigate the impact of the selection ratio on Text-Guided Patch Abstraction (TPA), we trained \modelname with varying selection ratios for the input token in the patch abstraction decoder of TPA. We kept the selection location of the Text Semantic-aware Patch Selector after the 6th transformer layer and the selection ratio for Key Patch Extraction at 70\% unchanged. As shown in Figure~\ref{fig:impact of input number}, two conclusions can be drawn: First, as the selection ratio increases, VQA performance first improves and then decreases, indicating that the “vanishing information" problem can hinder the effectiveness of the VLP model and highlighting the importance of shortening the visual path sequence. Second, setting the selection ratio to 20\% can achieves a good trade-off between effectiveness and efficiency.

% \begin{figure}[h]
% \centering
% \includegraphics[width=0.4\textwidth]{vqa_evit_pa.png}
% \caption{}
% \label{fig:MSMO}
% \end{figure}

\begin{table}[htpb!]
\centering
\footnotesize
\setlength{\tabcolsep}{1.6mm}{
\begin{tabular}{@{}ccccccc@{}}
\toprule[2.0pt]
 Loc         & SR-KPE &SR-TPA    & image size     & VQA & FLOPs & Troughout \\ \midrule
 - & - &  - &$384 \times 384$ &  77.55  & 84.87  & 63.57 \\ \hline
 6       & 70\%    & 20\% & $224\times224$ &     77.03        &\textbf{ 19.03} &  \textbf{ 360.23}   \\
  6       & 70\%     & 20\% & $256\times256$ &      77.61     &    24.04 &  255.03     \\
 6        & 70\%    & 20\% & $304\times304$ &    77.78         &   32.89  &  210.62\\
 6    & 70\%    & 20\% & $384\times384$ &      77.89       &    47.24 &  123.52   \\
6      & 70\%     & 20\% & $464\times464$ &     78.16     &  75.62  &  83.83  \\
6   & 70\%   & 20\% & $512\times512$ &     \textbf{78.28}    &  84.31 &  64.21   \\\bottomrule[2.0pt]
%  {[}5,10 {]}            & {[}70\%,70\%{]}& 512×512512\times512 &       76.36   & 83.07   &   100.02 & 61.81      \\ \bottomrule[2.0pt]
\end{tabular}}
\vspace{-2ex}
\centering
\caption{Results of \modelname finetuning on VQA task with different resolution images. The settings for calculating FLOPs and throughput are the same as Table \ref{table_efficient} except for the image resolution. The first row in the table reports the result of the recent SOTA baseline mPLUG\cite{li2022mplug}.}
\label{table: image_size}
\end{table}

\begin{table}[htpb!]
\centering
\footnotesize
\setlength{\tabcolsep}{1.2mm}{
\begin{tabular}{@{}lccccccc@{}}
\toprule[2.0pt]
model  &  Loc   & SR-KPE &SR-TPA    & VQA & FLOPs(G) & Throughput \\ \midrule
\modelname &  6       & 70\%    & 20\% &  77.89   & 19.03   &  360.23  \\ \hline
   \textit{-w/o KPE} &  -       & -    & 20\%  & 77.80 &   23.15 &     246.56  \\ 
   \textit{-w/o TPA} &  6       & 70\% & -  &77.64 &    25.63  &     232.42    \\ 
   \textit{-w/o BUS} &   -    & -   & - &77.58 &   37.62  &     170.23 \\ \hline
   \textit{-w/o IA} &  6       & 70\%    & 20\%   &77.67 &  18.89   &     362.40 \\
   \textit{-w/o TA} &  6       & 70\%    & 20\%   &77.43 &  18.67   &     366.23 \\
\bottomrule[2.0pt]
\end{tabular}}
\vspace{-2ex}
\caption{The result of ablations. We fine-tune \modelname on VQA and report test-dev results, FLOPs, and Throughput. The setting for calculating FLOPs and throughput is the same as Table~\ref{table_efficient}. }
\vspace{-3ex}
\label{table6}
\end{table}
\subsection{ The Impact of Selection Location and Selection Ratio for Key Patch Extraction }
\label{subsection:Impact_of_setting}
  To validate the impact of the location of the Text Semantic-aware Patch Selector (TSPS) in the ViT backbone and selected ratio in the KPE on the efficiency and effectiveness of \modelname, we trained \modelname with different selection locations and selection ratios. Note that when calculating FLOPs and throughput, we set the input image size to $224 \times 224$ and the input text length to 40. As shown in Table~\ref{table:impactofKR}, two conclusions can be drawn: (1), plugging the TSPS after shallower layers can reduce computational complexity but deteriorate accuracy. For example, the accuracy drops considerably with the remarkable increase in throughput when TSPS is placed after the 4th layer. A possible explanation is that patch embeddings in shallow layers cannot sufficiently represent visual semantics, making it difficult to learn fine-grained patch-text alignment, which leads to a drop in accuracy. (2), too many undetected image tokens fused in the TSPS module will considerably decrease downstream task performance. For example, if we place the TSPS after the 4th layer in ViT and set the keeping ratio to 40\%, performance will decrease to 76.22 on the VQA task, compared to 77.89 of the model with a 70\% selection ratio in the 6th layer.

\subsection{Fine-tuning on Higher Resolution Images}
\label{HigherImages}
% We can control the computational cost by fusing different numbers of inattentive tokens. Therefore, we finetune \modelname on the VQA task, which take images with varying resolutions as input. We report the results in Table                                                                                                                                  \ref{table: image_size}. The experimental results show that by increasing the input image resolution, we can facilitate the model by taking more image tokens to gain better performance. For example,  by finetuning \modelname with the images of 512×\times512, we can achieve the score of 78.25 on VQA, outperforming the baseline finetuned with images of 384×\times384 yet keeping similar computational complexity. 
   To test the efficacy of our approach, we fine-tune \modelname for the VQA task, taking images of varying resolutions as input. Table \ref{table: image_size} shows our results. Our experiments demonstrate that increasing the input image resolution facilitates the model by allowing it to take more image tokens, leading to improved performance. For instance, when fine-tuned with 512$\times$512 images, \modelname achieves a score of 78.28 on the VQA task, outperforming the baseline model fine-tuned with 384$\times$384 images while maintaining a similar level of computational complexity.
% \subsection{Effectiveness of Text-Relevant Image Patch Selection }
% To verify the effectiveness of Text-Relevant Image Patch Selection, we first implement the single-stream model \modelname-S. Then, we examine the downstream task performance, computational complexity, and inference speed of \modelname and \modelname-S (both with and without Text-Relevant Image Patch Selection). The results are shown in Table                                                                                                                                  ?\ref{table3}, and we find that for both \modelname and \modelname-S, we can see a consistent improvement in the inference speed and downstream task performance by incorporating the text-relevant image patch selection mechanism. These results suggest that the proposed image patch selection mechanism is not only efficient but also effective. Notably,  compared with the dual-stream model \modelname, \modelname-S is faster in inference due to the parameter efficiency of the single-stream model. However, its performance lags behind state-of-the-art performance on downstream VL tasks.

\subsection{Ablation Study}

  We conducted ablation studies to investigate the effects of our proposed bottom-up patch summarization mechanism. Specifically, we examined the impact of removing the Key Patch Extraction (KPE) and Text-Guided Patch Abstraction (TPA) on both performance and efficiency. In Table \ref{table6}, “w/o EPA" denotes the case where we remove KPE from the visual backbone but keep TPA outside the visual backbone, while “w/o TPA" is the opposite case where we remove TPA but keep KPE. “w/o BUS" indicates the removal of overall Bottom-Up Summarization from our VLP model. As shown in Table \ref{table6}, we observed that removing KPE and only using TPA on the top level did not result in a significant improvement in performance, but it led to a decrease in model efficiency. This suggests that the additional computational cost inside the ViT backbone cannot be ignored, and it highlights the effectiveness and efficiency of KPE. On the other hand, when we removed top-level TPA but kept bottom-level KPE, we observed a decrease in VQA performance and throughput, which indicates the importance of TPA in achieving good performance. Furthermore, when we removed both KPE and TPA (i.e., “w/o BUS"), we observed a more significant decrease in both performance and speed. This finding underscores the effectiveness and efficiency of our bottom-up patch summarization mechanism, which achieved a better trade-off between performance and efficiency.

 In addition, we also verified the effectiveness of the guidance of text semantics for patch summarization and the effectiveness of introducing attention maps of the visual [CLS] token in KPE. Specifically, “w/o TA" means that in KPE and TPA, we only use the attention maps of the visual [CLS] token to other tokens to select tokens. Table \ref{table6} shows a significant decrease in performance, which proves the effectiveness of the guidance of text semantics for patch summarization. In addition, “w/o IA" means that we only use the text alignment score predicted by TSPS to select tokens, without using visual information. We also found a slight decrease in performance, which confirms that introducing visual information (attention map of image [CLS] token) can address the bias introduced by bounding box labels mentioned in Subsection \ref{subsection:Key Patch Extraction }.

% The proposed Semantic-aware Patch Detection module detects the text-consistent image tokens in the vision backbone and preserves the detected image patches. To further investigate the effectiveness of the semantic-aware patch detection module, we visualize the VQA case and the detected text-relevant image patches in the Figure \ref{fig:casestudy}. It can be seen that based on the text questions, the SPD module can effectively detect the relevant patches, even only detecting 10\% patches, our model can still output the right answers. For examples, in the first case, the question is "Are they having breakfast?", the SPD can effectively detect the patches of food and girl's mouth in the image which are high relevant with the question. Note that these examples are not cherry-picked. The phenomenon in these two example is commonly observed among other samples. We demonstrate more cases in Appendix~\ref{sup:case study}.
\subsection{Case Study}
\begin{figure}
     \centering
     \includegraphics[width=0.45\textwidth]{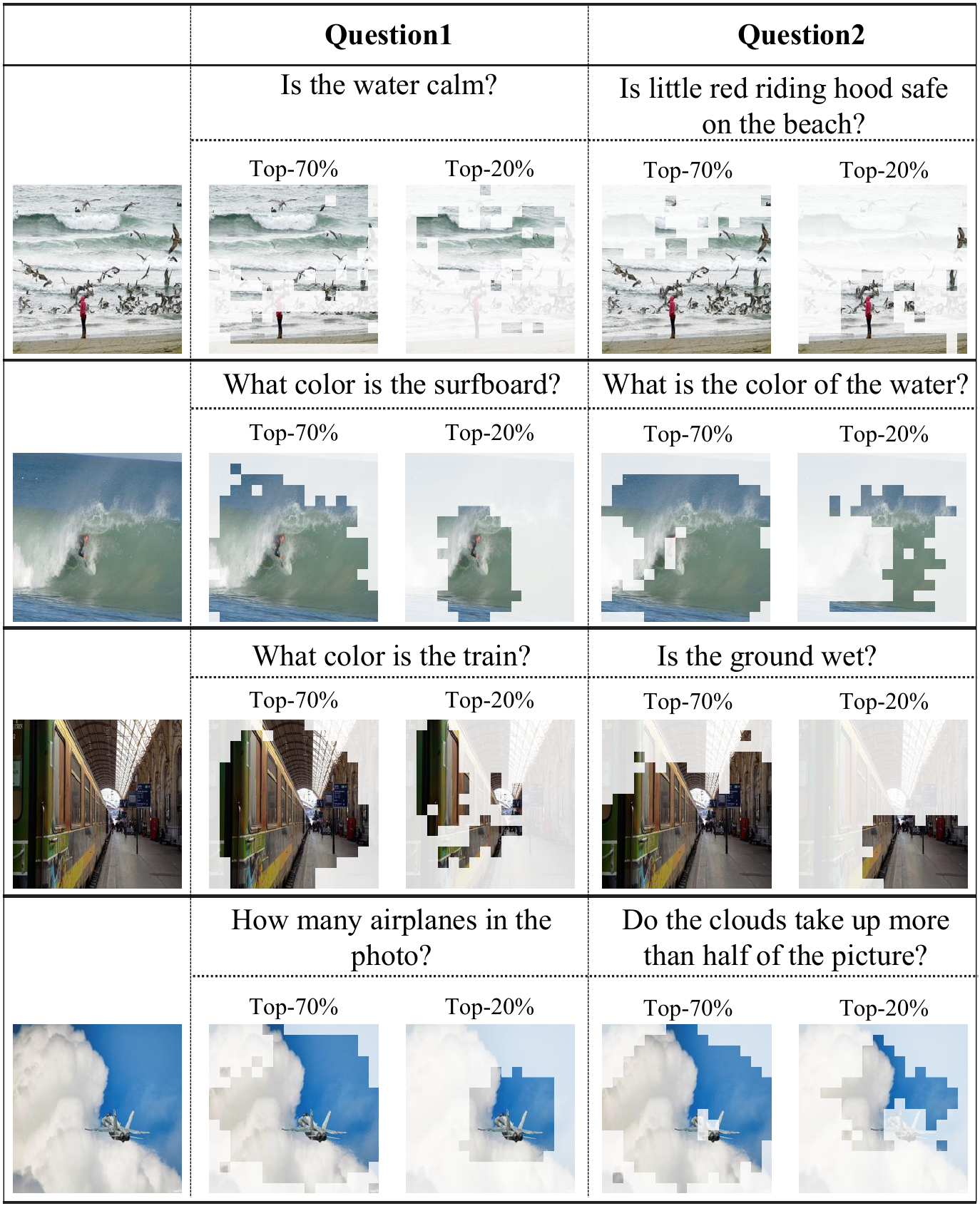}
     \caption{The visualization of the VQA case and the selected text-relevant image patches.}
     \label{fig:casestudy}
     \vspace{-3ex}
\end{figure}
  The proposed bottom-up patch summarization will select the text-consistent image tokens both in the ViT backbone and outside the ViT backbone with the regulation of the text semantics. To further investigate the effectiveness of text guidance for the bottom-up patch summarization, we visualize the VQA case and the selected text-relevant image patches in Figure \ref{fig:casestudy}. It can be seen that based on different text questions, we can effectively select the text-relevant patches while reducing the other text-irrelevant patches. For example, in the first case, the first question is “Is the water calm?", we can effectively select the patches of the water in the image while ignoring the human and the beach. For the second question “Is little red riding hood safe on the beach", we can still focus on the corresponding image patches (“little red riding hood" and “beach"). We demonstrate more cases in Appendix~\ref{sup:case study}.
\vspace{-2ex}
\section{Conclusion}

  We have presented \modelname with a novel Bottom-Up Patch Summarization mechanism that achieves an ideal trade-off between efficiency and effectiveness, resulting in a highly efficient and effective ViT-based VLP model. Our approach utilizes a Text-Semantics-Aware Patch Selector to perform bottom level Key Patch Extraction, followed by a Transformer-based Patch Abstraction Decoder for top-level visual abstraction. The combination of the two components enables our \modelname to learn a concise summary of lengthy visual token sequences efficiently and effectively. The experiment shows our method improves efficiency due to the reduction of visual sequences while keeping or even improving the performance of downstream tasks.
{\small
\bibliographystyle{ieee_fullname}
\bibliography{egbib}

\begin{thebibliography}{10}\itemsep=-1pt

\bibitem{Agrawal2015VQAVQ}
Aishwarya Agrawal, Jiasen Lu, Stanislaw Antol, Margaret Mitchell, C.~Lawrence Zitnick, Devi Parikh, and Dhruv Batra.
\newblock Vqa: Visual question answering.
\newblock {\em International Journal of Computer Vision}, 123:4--31, 2015.

\bibitem{nocaps}
Harsh Agrawal, Karan Desai, Yufei Wang, Xinlei Chen, Rishabh Jain, Mark Johnson, Dhruv Batra, Devi Parikh, Stefan Lee, and Peter Anderson.
\newblock nocaps: novel object captioning at scale.
\newblock {\em CoRR}, abs/1812.08658, 2018.

\bibitem{Alayrac2022FlamingoAV}
Jean-Baptiste Alayrac, Jeff Donahue, Pauline Luc, Antoine Miech, Iain Barr, Yana Hasson, Karel Lenc, Arthur Mensch, Katie Millican, Malcolm Reynolds, Roman Ring, Eliza Rutherford, Serkan Cabi, Tengda Han, Zhitao Gong, Sina Samangooei, Marianne Monteiro, Jacob Menick, Sebastian Borgeaud, Andy Brock, Aida Nematzadeh, Sahand Sharifzadeh, Mikolaj Binkowski, Ricardo Barreira, Oriol Vinyals, Andrew Zisserman, and Karen Simonyan.
\newblock Flamingo: a visual language model for few-shot learning.
\newblock {\em ArXiv}, abs/2204.14198, 2022.

\bibitem{anderson1988teaching}
Valerie Anderson and Suzanne Hidi.
\newblock Teaching students to summarize.
\newblock {\em Educational leadership}, 46(4):26--28, 1988.

\bibitem{bi2020palm}
Bin Bi, Chenliang Li, Chen Wu, Ming Yan, Wei Wang, Songfang Huang, Fei Huang, and Luo Si.
\newblock Palm: Pre-training an autoencoding\&autoregressive language model for context-conditioned generation.
\newblock {\em arXiv preprint arXiv:2004.07159}, 2020.

\bibitem{Caron2021EmergingPI}
Mathilde Caron, Hugo Touvron, Ishan Misra, Herv'e J'egou, Julien Mairal, Piotr Bojanowski, and Armand Joulin.
\newblock Emerging properties in self-supervised vision transformers.
\newblock {\em 2021 IEEE/CVF International Conference on Computer Vision (ICCV)}, pages 9630--9640, 2021.

\bibitem{Chen2020UNITERUI}
Yen-Chun Chen, Linjie Li, Licheng Yu, Ahmed~El Kholy, Faisal Ahmed, Zhe Gan, Yu Cheng, and Jingjing Liu.
\newblock Uniter: Universal image-text representation learning.
\newblock In {\em ECCV}, 2020.

\bibitem{Cheng2016NeuralSB}
Jianpeng Cheng and Mirella Lapata.
\newblock Neural summarization by extracting sentences and words.
\newblock {\em ArXiv}, abs/1603.07252, 2016.

\bibitem{cubuk2020randaugment}
Ekin~D Cubuk, Barret Zoph, Jonathon Shlens, and Quoc~V Le.
\newblock Randaugment: Practical automated data augmentation with a reduced search space.
\newblock In {\em Proceedings of the IEEE/CVF Conference on Computer Vision and Pattern Recognition Workshops}, pages 702--703, 2020.

\bibitem{Devlin2019BERTPO}
Jacob Devlin, Ming-Wei Chang, Kenton Lee, and Kristina Toutanova.
\newblock Bert: Pre-training of deep bidirectional transformers for language understanding.
\newblock {\em ArXiv}, abs/1810.04805, 2019.

\bibitem{dou2021empirical}
Zi-Yi Dou, Yichong Xu, Zhe Gan, Jianfeng Wang, Shuohang Wang, Lijuan Wang, Chenguang Zhu, Zicheng Liu, Michael Zeng, et~al.
\newblock An empirical study of training end-to-end vision-and-language transformers.
\newblock {\em arXiv preprint arXiv:2111.02387}, 2021.

\bibitem{gan2020large}
Zhe Gan, Yen-Chun Chen, Linjie Li, Chen Zhu, Yu Cheng, and Jingjing Liu.
\newblock Large-scale adversarial training for vision-and-language representation learning.
\newblock In {\em NeurIPS}, 2020.

\bibitem{goyal2017making}
Yash Goyal, Tejas Khot, Douglas Summers-Stay, Dhruv Batra, and Devi Parikh.
\newblock Making the v in vqa matter: Elevating the role of image understanding in visual question answering.
\newblock In {\em Proceedings of the IEEE conference on computer vision and pattern recognition}, pages 6904--6913, 2017.

\bibitem{Huang2020PixelBERTAI}
Zhicheng Huang, Zhaoyang Zeng, Bei Liu, Dongmei Fu, and Jianlong Fu.
\newblock Pixel-bert: Aligning image pixels with text by deep multi-modal transformers.
\newblock {\em ArXiv}, abs/2004.00849, 2020.

\bibitem{jia2021scaling}
Chao Jia, Yinfei Yang, Ye Xia, Yi-Ting Chen, Zarana Parekh, Hieu Pham, Quoc~V Le, Yunhsuan Sung, Zhen Li, and Tom Duerig.
\newblock Scaling up visual and vision-language representation learning with noisy text supervision.
\newblock {\em arXiv preprint arXiv:2102.05918}, 2021.

\bibitem{Jiang2023ExploitingPI}
Chaoya Jiang, Rui Xie, Wei Ye, Jinan Sun, and Shikun Zhang.
\newblock Exploiting pseudo image captions for multimodal summarization.
\newblock In {\em Annual Meeting of the Association for Computational Linguistics}, 2023.

\bibitem{Jiang2022TRIPSEV}
Chaoya Jiang, Haiyang Xu, Chenliang Li, Ming Yan, Wei Ye, Shikun Zhang, Bin Bi, and Songfang Huang.
\newblock Trips: Efficient vision-and-language pre-training with text-relevant image patch selection.
\newblock In {\em Conference on Empirical Methods in Natural Language Processing}, 2022.

\bibitem{Jiang2023COPAE}
Chaoya Jiang, Haiyang Xu, Wei Ye, Qinghao Ye, Chenliang Li, Mingshi Yan, Bin Bi, Shikun Zhang, Ji Zhang, and Feiyan Huang.
\newblock Copa : Efficient vision-language pre-training through collaborative object- and patch-text alignment.
\newblock {\em Proceedings of the 31st ACM International Conference on Multimedia}, 2023.

\bibitem{Jiang2023VisionLP}
Chaoya Jiang, Wei Ye, Haiyang Xu, Miang yan, Shikun Zhang, Jie Zhang, and Fei Huang.
\newblock Vision language pre-training by contrastive learning with cross-modal similarity regulation.
\newblock In {\em Annual Meeting of the Association for Computational Linguistics}, 2023.

\bibitem{jing1999decomposition}
Hongyan Jing and Kathleen~R McKeown.
\newblock The decomposition of human-written summary sentences.
\newblock In {\em Proceedings of the 22nd annual international ACM SIGIR conference on Research and development in information retrieval}, pages 129--136, 1999.

\bibitem{Kamath2021MDETRM}
Aishwarya Kamath, Mannat Singh, Yann LeCun, Ishan Misra, Gabriel Synnaeve, and Nicolas Carion.
\newblock Mdetr - modulated detection for end-to-end multi-modal understanding.
\newblock {\em 2021 IEEE/CVF International Conference on Computer Vision (ICCV)}, pages 1760--1770, 2021.

\bibitem{karpathy2015deep}
Andrej Karpathy and Li Fei-Fei.
\newblock Deep visual-semantic alignments for generating image descriptions.
\newblock In {\em Proceedings of the IEEE conference on computer vision and pattern recognition}, pages 3128--3137, 2015.

\bibitem{Kim2021ViLTVT}
Wonjae Kim, Bokyung Son, and Ildoo Kim.
\newblock Vilt: Vision-and-language transformer without convolution or region supervision.
\newblock In {\em ICML}, 2021.

\bibitem{Krishna2016VisualGC}
Ranjay Krishna, Yuke Zhu, Oliver Groth, Justin Johnson, Kenji Hata, Joshua Kravitz, Stephanie Chen, Yannis Kalantidis, Li-Jia Li, David~A. Shamma, Michael~S. Bernstein, and Li Fei-Fei.
\newblock Visual genome: Connecting language and vision using crowdsourced dense image annotations.
\newblock {\em International Journal of Computer Vision}, 123:32--73, 2016.

\bibitem{Lewis2020BARTDS}
Mike Lewis, Yinhan Liu, Naman Goyal, Marjan Ghazvininejad, Abdelrahman Mohamed, Omer Levy, Veselin Stoyanov, and Luke Zettlemoyer.
\newblock Bart: Denoising sequence-to-sequence pre-training for natural language generation, translation, and comprehension.
\newblock {\em ArXiv}, abs/1910.13461, 2020.

\bibitem{li2022mplug}
Chenliang Li, Haiyang Xu, Junfeng Tian, Wei Wang, Ming Yan, Bin Bi, Jiabo Ye, Hehong Chen, Guohai Xu, Zheng Cao, Ji Zhang, Songfang Huang, Fei Huang, Jingren Zhou, and Luo Si.
\newblock mplug: Effective and efficient vision-language learning by cross-modal skip-connections, 2022.

\bibitem{Li2023BLIP2BL}
Junnan Li, Dongxu Li, Silvio Savarese, and Steven Hoi.
\newblock Blip-2: Bootstrapping language-image pre-training with frozen image encoders and large language models.
\newblock {\em ArXiv}, abs/2301.12597, 2023.

\bibitem{li2022blip}
Junnan Li, Dongxu Li, Caiming Xiong, and Steven Hoi.
\newblock Blip: Bootstrapping language-image pre-training for unified vision-language understanding and generation.
\newblock {\em arXiv preprint arXiv:2201.12086}, 2022.

\bibitem{Li2021AlignBF}
Junnan Li, Ramprasaath~R. Selvaraju, Akhilesh~Deepak Gotmare, Shafiq~R. Joty, Caiming Xiong, and Steven C.~H. Hoi.
\newblock Align before fuse: Vision and language representation learning with momentum distillation.
\newblock In {\em NeurIPS}, 2021.

\bibitem{Li2019VisualBERTAS}
Liunian~Harold Li, Mark Yatskar, Da Yin, Cho-Jui Hsieh, and Kai-Wei Chang.
\newblock Visualbert: A simple and performant baseline for vision and language.
\newblock {\em ArXiv}, abs/1908.03557, 2019.

\bibitem{Li2020OscarOA}
Xiujun Li, Xi Yin, Chunyuan Li, Xiaowei Hu, Pengchuan Zhang, Lei Zhang, Lijuan Wang, Houdong Hu, Li Dong, Furu Wei, Yejin Choi, and Jianfeng Gao.
\newblock Oscar: Object-semantics aligned pre-training for vision-language tasks.
\newblock In {\em ECCV}, 2020.

\bibitem{Liang2022EViTEV}
Youwei Liang, Chongjian Ge, Zhan Tong, Yibing Song, Jue Wang, and Pengtao Xie.
\newblock Evit: Expediting vision transformers via token reorganizations.
\newblock In {\em International Conference on Learning Representations}, 2022.

\bibitem{Lin2014MicrosoftCC}
Tsung-Yi Lin, Michael Maire, Serge~J. Belongie, James Hays, Pietro Perona, Deva Ramanan, Piotr Doll{\'a}r, and C.~Lawrence Zitnick.
\newblock Microsoft coco: Common objects in context.
\newblock In {\em ECCV}, 2014.

\bibitem{Loshchilov2019DecoupledWD}
Ilya Loshchilov and Frank Hutter.
\newblock Decoupled weight decay regularization.
\newblock In {\em ICLR}, 2019.

\bibitem{Lu2019ViLBERTPT}
Jiasen Lu, Dhruv Batra, Devi Parikh, and Stefan Lee.
\newblock Vilbert: Pretraining task-agnostic visiolinguistic representations for vision-and-language tasks.
\newblock In {\em NeurIPS}, 2019.

\bibitem{nallapati2017summarunner}
Ramesh Nallapati, Feifei Zhai, and Bowen Zhou.
\newblock Summarunner: A recurrent neural network based sequence model for extractive summarization of documents.
\newblock In {\em Thirty-First AAAI Conference on Artificial Intelligence}, 2017.

\bibitem{narayan2018ranking}
Shashi Narayan, Shay~B Cohen, and Mirella Lapata.
\newblock Ranking sentences for extractive summarization with reinforcement learning.
\newblock In {\em Proceedings of the 2018 Conference of the North American Chapter of the Association for Computational Linguistics: Human Language Technologies, Volume 1 (Long Papers)}, pages 1747--1759, 2018.

\bibitem{Ordonez2011Im2TextDI}
Vicente Ordonez, Girish Kulkarni, and Tamara~L. Berg.
\newblock Im2text: Describing images using 1 million captioned photographs.
\newblock In {\em NIPS}, 2011.

\bibitem{paulus2018deep}
Romain Paulus, Caiming Xiong, and Richard Socher.
\newblock A deep reinforced model for abstractive summarization.
\newblock In {\em International Conference on Learning Representations}, 2018.

\bibitem{Plummer2015Flickr30kEC}
Bryan~A. Plummer, Liwei Wang, Christopher~M. Cervantes, Juan~C. Caicedo, J. Hockenmaier, and Svetlana Lazebnik.
\newblock Flickr30k entities: Collecting region-to-phrase correspondences for richer image-to-sentence models.
\newblock {\em International Journal of Computer Vision}, 123:74--93, 2015.

\bibitem{Radford2021LearningTV}
Alec Radford, Jong~Wook Kim, Chris Hallacy, Aditya Ramesh, Gabriel Goh, Sandhini Agarwal, Girish Sastry, Amanda Askell, Pamela Mishkin, Jack Clark, Gretchen Krueger, and Ilya Sutskever.
\newblock Learning transferable visual models from natural language supervision.
\newblock In {\em ICML}, 2021.

\bibitem{Rao2021DynamicViTEV}
Yongming Rao, Wenliang Zhao, Benlin Liu, Jiwen Lu, Jie Zhou, and Cho-Jui Hsieh.
\newblock Dynamicvit: Efficient vision transformers with dynamic token sparsification.
\newblock In {\em NeurIPS}, 2021.

\bibitem{scst}
Steven~J. Rennie, Etienne Marcheret, Youssef Mroueh, Jerret Ross, and Vaibhava Goel.
\newblock Self-critical sequence training for image captioning.
\newblock In {\em 2017 IEEE Conference on Computer Vision and Pattern Recognition (CVPR)}, pages 1179--1195, 2017.

\bibitem{see2017get}
Abigail See, Peter~J Liu, and Christopher~D Manning.
\newblock Get to the point: Summarization with pointer-generator networks.
\newblock In {\em Proceedings of the 55th Annual Meeting of the Association for Computational Linguistics (Volume 1: Long Papers)}, pages 1073--1083, 2017.

\bibitem{sharma2018conceptualCA}
Piyush Sharma, Nan Ding, Sebastian Goodman, and Radu Soricut.
\newblock Conceptual captions: A cleaned, hypernymed, image alt-text dataset for automatic image captioning.
\newblock In {\em ACL}, 2018.

\bibitem{Su2020VLBERTPO}
Weijie Su, Xizhou Zhu, Yue Cao, Bin Li, Lewei Lu, Furu Wei, and Jifeng Dai.
\newblock Vl-bert: Pre-training of generic visual-linguistic representations.
\newblock {\em ArXiv}, abs/1908.08530, 2020.

\bibitem{Syed2021ASO}
Ayesha~Ayub Syed, Ford~Lumban Gaol, and Tokuro Matsuo.
\newblock A survey of the state-of-the-art models in neural abstractive text summarization.
\newblock {\em IEEE Access}, 9:13248--13265, 2021.

\bibitem{Tan2019LXMERTLC}
Hao~Hao Tan and Mohit Bansal.
\newblock Lxmert: Learning cross-modality encoder representations from transformers.
\newblock {\em ArXiv}, abs/1908.07490, 2019.

\bibitem{Tan2017AbstractiveDS}
Jiwei Tan, Xiaojun Wan, and Jianguo Xiao.
\newblock Abstractive document summarization with a graph-based attentional neural model.
\newblock In {\em ACL}, 2017.

\bibitem{Wang2020MiniVLMAS}
Jianfeng Wang, Xiaowei Hu, Pengchuan Zhang, Xiujun Li, Lijuan Wang, L. Zhang, Jianfeng Gao, and Zicheng Liu.
\newblock Minivlm: A smaller and faster vision-language model.
\newblock {\em ArXiv}, abs/2012.06946, 2020.

\bibitem{Wang2021VLMoUV}
Wenhui Wang, Hangbo Bao, Li Dong, and Furu Wei.
\newblock Vlmo: Unified vision-language pre-training with mixture-of-modality-experts.
\newblock {\em ArXiv}, abs/2111.02358, 2021.

\bibitem{Wang2021SimVLMSV}
Zirui Wang, Jiahui Yu, Adams~Wei Yu, Zihang Dai, Yulia Tsvetkov, and Yuan Cao.
\newblock Simvlm: Simple visual language model pretraining with weak supervision.
\newblock {\em ArXiv}, abs/2108.10904, 2021.

\bibitem{xiao2019extractive}
Wen Xiao and Giuseppe Carenini.
\newblock Extractive summarization of long documents by combining global and local context.
\newblock In {\em Proceedings of the 2019 Conference on Empirical Methods in Natural Language Processing and the 9th International Joint Conference on Natural Language Processing (EMNLP-IJCNLP)}, pages 3011--3021, 2019.

\bibitem{Xu2021E2EVLPEV}
Haiyang Xu, Ming Yan, Chenliang Li, Bin Bi, Songfang Huang, Wenming Xiao, and Fei Huang.
\newblock E2e-vlp: End-to-end vision-language pre-training enhanced by visual learning.
\newblock {\em ArXiv}, abs/2106.01804, 2021.

\bibitem{Xu2023mPLUG2AM}
Haiyang Xu, Qinghao Ye, Mingshi Yan, Yaya Shi, Jiabo Ye, Yuanhong Xu, Chenliang Li, Bin Bi, Qiuchen Qian, Wei Wang, Guohai Xu, Ji Zhang, Songfang Huang, Feiran Huang, and Jingren Zhou.
\newblock mplug-2: A modularized multi-modal foundation model across text, image and video.
\newblock {\em ArXiv}, abs/2302.00402, 2023.

\bibitem{Yang2021CrossingTF}
Zhengyuan Yang, Zhe Gan, Jianfeng Wang, Xiaowei Hu, Faisal Ahmed, Zicheng Liu, Yumao Lu, and Lijuan Wang.
\newblock Crossing the format boundary of text and boxes: Towards unified vision-language modeling.
\newblock {\em ArXiv}, abs/2111.12085, 2021.

\bibitem{DBLP:journals/corr/abs-2111-12085}
Zhengyuan Yang, Zhe Gan, Jianfeng Wang, Xiaowei Hu, Faisal Ahmed, Zicheng Liu, Yumao Lu, and Lijuan Wang.
\newblock Crossing the format boundary of text and boxes: Towards unified vision-language modeling.
\newblock {\em CoRR}, abs/2111.12085, 2021.

\bibitem{Ye2023mPLUGOwlME}
Qinghao Ye, Haiyang Xu, Guohai Xu, Jiabo Ye, Ming Yan, Yi Zhou, Junyan Wang, Anwen Hu, Pengcheng Shi, Yaya Shi, Chenliang Li, Yuanhong Xu, Hehong Chen, Junfeng Tian, Qiang Qi, Ji~Chao Zhang, and Feiyan Huang.
\newblock mplug-owl: Modularization empowers large language models with multimodality.
\newblock {\em ArXiv}, abs/2304.14178, 2023.

\bibitem{Yu2021ERNIEViLKE}
Fei Yu, Jiji Tang, Weichong Yin, Yu Sun, Hao Tian, Hua Wu, and Haifeng Wang.
\newblock Ernie-vil: Knowledge enhanced vision-language representations through scene graph.
\newblock In {\em AAAI}, 2021.

\bibitem{yu2016modeling}
Licheng Yu, Patrick Poirson, Shan Yang, Alexander~C Berg, and Tamara~L Berg.
\newblock Modeling context in referring expressions.
\newblock In {\em European Conference on Computer Vision}, pages 69--85. Springer, 2016.

\bibitem{Zeng2021xvlm}
Yan Zeng, Xinsong Zhang, and Hang Li.
\newblock Multi-grained vision language pre-training: Aligning texts with visual concepts.
\newblock {\em ArXiv}, abs/2111.08276, 2021.

\bibitem{zhang2020pegasus}
Jingqing Zhang, Yao Zhao, Mohammad Saleh, and Peter Liu.
\newblock Pegasus: Pre-training with extracted gap-sentences for abstractive summarization.
\newblock In {\em International Conference on Machine Learning}, pages 11328--11339. PMLR, 2020.

\bibitem{2021VinVL}
P. Zhang, X. Li, X. Hu, J. Yang, L. Zhang, L. Wang, Y. Choi, and J. Gao.
\newblock Vinvl: Making visual representations matter in vision-language models.
\newblock 2021.

\bibitem{zhong2020extractive}
Ming Zhong, Pengfei Liu, Yiran Chen, Danqing Wang, Xipeng Qiu, and Xuan-Jing Huang.
\newblock Extractive summarization as text matching.
\newblock In {\em Proceedings of the 58th Annual Meeting of the Association for Computational Linguistics}, pages 6197--6208, 2020.

\end{thebibliography}
}
\clearpage
\appendix

\section{Pretraining details}
\label{sup:pretraining detail}

We employed the AdamW optimizer~\cite{Loshchilov2019DecoupledWD} with a weight decay of 0.02 for training our models. The learning rate was warmed up to 1e-5 (ViT-B/16) and 1e-4 (BERT$_{base}$) in the first 1000 iterations, then decayed to 1e-6 according to a cosine schedule. The pre-training of \modelname required approximately 60 hours and was performed on 8 A100-80G GPUs using a 4M pre-training dataset for about 20 epochs.

To improve the generalization of vision encoders during pre-training, we applied RandAugment~\cite{cubuk2020randaugment} to random image crops of size 256 $\times$ 256. In the fine-tuning stage for VQA, image captioning, and visual grounding tasks, we increased the image resolution. For image-text contrastive learning, we set the queue size to 65,536 and the momentum coefficient to 0.995.

\subsection{Pretraining data}
\vspace{-2ex}
\label{sup:pretraining data}
\begin{table}[htbp]
\setlength\tabcolsep{12pt}
\centering
\small
\begin{tabular}{l|cccc}
\toprule[1.5pt]
  &  COCO & VG & SBU & CC3M \\
\midrule
image & 113K & 100K & 860K & 3M   \\
text & 567K & 769K & 860K & 3M \\
\bottomrule[1.5pt]
\end{tabular} 
\caption{Statistics of the pre-training datasets.}
\label{table:pretraindata}
\vspace{-3ex}
\end{table}

\begin{table}[htbp]
\setlength\tabcolsep{10pt}
\centering
\small
\begin{tabular}{l|cccc}
\toprule[1.5pt]
     &  image & Captions & Objects & Regions \\
\midrule
COCO & 0.11M & 0.55M & 0.45M & -   \\
VG   & 0.10M & - & 2.0M & 3.7M \\
\bottomrule[1.5pt]
\end{tabular} 
\caption{Statistics of objects/regions annotations used in the pre-training.}
\label{table:objectdata}
\vspace{-2ex}

\end{table}
% Table \ref{table:pretraindata} shows the statistics of the 4M images with texts used in the pre-training stage. Besides, As shown in Table~\ref{table:objectdata} we use also use the objects/regions annotions from COCO\cite{Lin2014MicrosoftCC} and VG \cite{Krishna2016VisualGC} datasets and we give a statistics of object and region annotations of each dataset. Note that we use the object/region annotations provide by ~\cite{Zeng2021xvlm} thus we following their setting which filtered out some samples because of: 1) invalid annotations (e.g. negative values for bounding boxes or boxes being outside of the images); 2) boxes being too small (< 1\%); 3) highly overlapped textual descriptions of regions (>75\%), etc. After pre-processing, we keep: for example, COCO objects 446,873 (from 859,999), VG objects 2,043,927 (from 3,802,349), VG regions 3,699,598 (from 5,402,953).
Table \ref{table:pretraindata} shows the statistics of the 4M images with texts used in the pre-training stage. Additionally, we use object/region annotations from the COCO~\cite{Lin2014MicrosoftCC} and VG~\cite{Krishna2016VisualGC} datasets, as shown in Table~\ref{table:objectdata}, and provide statistics of object and region annotations for each dataset. We follow the object/region annotations provided by~\cite{Zeng2021xvlm}, which filter out some samples due to: 1) invalid annotations (e.g., negative values for bounding boxes or boxes being outside of the images); 2) boxes being too small ($\leq$ 1\%); 3) highly overlapped textual descriptions of regions ( $\geq$ 75\%), etc. After pre-processing, we keep 446,873 COCO objects (from 859,999), 2,043,927 VG objects (from 3,802,349), and 3,699,598 VG regions (from 5,402,953).
\subsection{Pretraining Task}
\label{sup:pretraining Task}
We pre-train our model with five standard objectives: Image-Text Contrastive learning (ITC), Image-Text Matching (ITM), and  Masked Language Modeling (MLM),Prefix Language Modeling (PrefixLM), \PretrainTaskName (PTM). These pre-training tasks are optimized jointly. In this subsection, we will firstly introduce the last four pre-training task and then give the details of the \PretrainTaskName. 

\noindent \textbf{Image-text Contrastive (ITC)} For \modelname, We follow the \cite{Li2021AlignBF} and apply ITC to align the image representation and text representation from the unimodal encoders. For the image, the image feature corresponding to the image [CLS] token is chosen as the image representation. For the text, the text token feature corresponding to the text [CLS] token is the text representation.

\noindent \textbf{Image-Text Matching (ITM)}  The goal of image-text matching is to predict whether the input image and text are matched.  We follow the design of \cite{Li2021AlignBF} and select hard negative image-text pairs based on the contrastive text-image similarity. We take the text [CLS] embedding of the multimodal encoder's output as the joint representation, followed by a Multi-Layer Perceptron (MLP) layer for prediction.

\noindent \textbf{Masked Language Modeling (MLM)} The task setup is basically the same as in BERT~\cite{Devlin2019BERTPO}, where we randomly mask 15$\%$ of tokens in text and the model is asked to predict these masked words with the cross-modal representations.

\noindent \textbf{Prefix Language Modeling (PrefixLM).} This task aims to generate the caption given an image and predict the text segment subsequent to the cross-modal context as ~\cite{bi2020palm}. It optimizes a cross entropy loss by maximizing the likelihood of text in an autoregressive manner.

\subsubsection{Patch-Text Matching}

The key component for the bottom-up patch summarization is the Text Semantic-aware Patch Selector (TSPS) which needs to predict the fine-grained alignment scores between the image patches and input text to select the text-relevant patches. However, such fine-grained patch-text alignment capabilities of traditional ViT-based models are weak as the lack of fine-grained patch-text labels. To address the above difficulties, we introduce a novel pre-training task named Patch Text Matching (PTM) which facilitates the patch detector training and drives our model to learn the fine-grained patch-text alignment.

In most object objection and visual grounding datasets, objects and regions are typically paired with a class label or text description. Therefore, for each (object/region) bounding box in an image, we can obtain a corresponding text description (For the object class label, we can transfer it to a text description using a text template such as “this is a [Class Label]"). We then transform the bounding box annotations into patch-level labels by assigning a label of 1 to an image patch if it overlaps with the bounding box and 0 otherwise. Different text descriptions and bounding boxes result in different patch labels, enabling us to generate fine-grained patch-text labels that serve as supervisory signals for pre-training our model.

During pre-training, we randomly sample a mini-batch of images from object detection/visual grounding datasets such as COCO~\cite{Lin2014MicrosoftCC} or VG~\cite{Krishna2016VisualGC}. For each image, we randomly select an object/region bounding box and translate the bounding box annotation to the image patch label sequence following the aforementioned transformation rule. We then feed the batch of text descriptions of the bounding boxes and the images to BUS. We expect the TSPS to predict all patches that overlap with the bounding box with the guidance of the bounding box text description.

Once TSPS has predicted the alignment scores between image patches and text, we calculate the binary cross-entropy loss between the alignment scores and patch labels using the following equation:
\vspace{-1ex}
\begin{equation}
    \mathbf{L}_{PTM} = \frac{1}{n}\sum_{i=1}^{n} Y_{i} log\left(a_i\right) + \left(1-Y_{i}\right)log\left(1-a_i\right) 
\end{equation}

Here, $a_i$ is the alignment score between the $i^{th}$ patch in the image and the input text, and $Y_i$ is the patch label of the $i^{th}$ patch. 
Besides, at the beginning of pre-training, as the PTM loss has not yet converged, thus the performance of the patch selector is not ideal, we select the image patches directly based on the attention weights of the image [CLS] token to other patch tokens by setting the hyper-parameter $\beta$ to 0. As the PTM loss gradually converges, we will progressively set a large value to $\beta$.

After calculating the PTM loss $\mathbf{L}_{PTM}$, we then randomly sample a mini-batch of normal image-text pairs from the dataset of 4M images  and calculate the Image-Text Contrastive (ITC) loss $\mathbf{L}_{ITC}$, Image-Text Matching (ITM) loss $\mathbf{L}_{ITM}$, Masked Language Modeling (MLM) loss $\mathbf{L}_{MLM}$ and Prefix Language Modeling (PrefixLM) loss $\mathbf{L}_{Prefix}$ based on other four pre-training objectives. We assign equal loss weights to each pre-training loss, and thus the full pre-training loss is:

\begin{equation}
     \mathbf{L} = \mathbf{L}_{ITC}+\mathbf{L}_{ITM} +\mathbf{L}_{MLM} + \mathbf{L}_{Prefix} + \mathbf{L}_{PTM}
\end{equation}

\subsection{Pretraining Schedule}
\label{sup:pretraining schedule}
In this subsection, as shown in Algorithm \ref{alg:pretrain}, we give a algorithm of the pretraining schedule of our model \modelname.
\vspace{-1ex}
\begin{algorithm}[htbp]
  \caption{Pre-training of BUS}
  \label{alg:pretrain}

  \KwIn{Large scale pretraining dataset $\mathcal{D}$, Object/Region Dataset $\mathcal{O}$, the number of pre-training epochs $T$, the pre-training learning rate $\alpha$, the batch size $B_D$ of dataset $\mathcal{D}$, the batch size $B_O$ of dataset $\mathcal{O}$.  }

  Initialize the parameters $\theta$ of our model $M$ \;
  \For{$t=1$ to $T$}{
    Randomly sample a mini-batch of $B_O$ Images $\{\hat{v}_1, \hat{v}_2, \dots, \hat{v}_{B_O} \}$ from $\mathcal{D}$ \;
    \For{$i=1$ to $B_O$}{
      Select a object or region $r_i$ from image $\hat{v}_i$ \;
      Translate the object class label $\hat{y}_i$ to text description $\hat{t}_i$\;
      Translate the bounding box annotation of $r_i$ to patch annotations $ Y^i = \{y^i_1,y^i_2, \dots, y^i_n\}$ \;
     
    }
    Run forward of $M$ on the mini-batch of image-text pairs $\{\{\hat{v}_1, \hat{t}_1\}, \{\hat{v}_2, \hat{t}_2\}, \dots, \{\hat{v}_{B_O}, \hat{t}_{B_O}\} \}$ and $\{Y^1, Y^2, \dots, Y^{B_O}\}$ to obtain the loss $\mathcal{L}_{PTM}$ \;

    Randomly sample a mini-batch of $B$ Image-Text Pairs $\{\{v_1,t_1\}, \{v_2,t_3\}, \ldots, \{v_{B_D},t_{B_D}\}\}$ from $\mathcal{D}$ \;

    Run forward of $M$ on the mini-batch of image-text pairs $\{\{v_1,t_1\}, \{v_2,t_3\}, \ldots, \{v_{B_D},t_{B_D}\}\}$  to obtain the losses $\mathcal{L}_{ITC}$, $\mathcal{L}_{ITM}$, $\mathcal{L}_{MLM}$, $\mathcal{L}_{Prefix}$ \;
    
    Calculate the overall loss:
    
    $\mathbf{L} = \mathbf{L}_{ITC}+\mathbf{L}_{ITM} +\mathbf{L}_{MLM} + \mathbf{L}_{Prefix} + \mathbf{L}_{PTM}$\;
    
    Backward the overall loss $\mathbf{L}$ and update the parameters of $M$ using gradient descent with learning rate $\alpha$ and the average loss $\mathbf{L}$ over the mini-batch:

    $\theta \leftarrow \theta - \alpha \frac{1}{B} \sum_{i=1}^{B} \nabla_{\theta} \mathcal{L}(\theta; s_i)$ \;

  }

  \Return $M$ with pre-trained parameters $\theta$ \;

\end{algorithm}

\section{Downstream Task Details}
\label{sup:downstream task details}

We evaluate \modelname on the four downstream vision-language tasks. The hyperparameters that we use for finetuning on the downstream tasks are listed in Table \ref{table:finetune-hyper}. Following ~\cite{Li2021AlignBF}, all tasks adopt RandAugment, AdamW optimizer with a weight decay of 0.05 and a cosine learning rate schedule. Next we introduce the dataset settings in detail.

% \subsection{Ablation Studies of Pre-training Objectives}

\paragraph{VQA.} The VQA task ~\cite{Agrawal2015VQAVQ} requires the model to answer natural language questions given an image. Most methods~\cite{Tan2019LXMERTLC,Wang2021VLMoUV,Li2020OscarOA,Wang2021SimVLMSV} deal with visual question answering tasks as multi-label classification on pre-defined answer sets. This strategy achieves strong performance, but it is not suitable for real-world open scenarios. We conduct experiment on the VQA2.0 dataset~\cite{goyal2017making}, which contains 83k/41k/81k images for training/validation/test. Following ~\cite{Li2021AlignBF}, we use both training and validation splits for training, and incorporate additional training data from Visual Genome~\cite{Krishna2016VisualGC}. Following \cite{Li2020OscarOA}, we concatenate the question with the object labels and OCR tokens extracted from image.
% Moreover, we directly select the silent image according to the ROUGE-L scores between the aligned sentences or image caption
\begin{table}
\setlength\tabcolsep{4pt}
\centering
\small
\begin{tabular}{l|ccc}
\toprule[1.5pt]
Task  &  LR (ViT-L/BERT$_{base}$) & batch size & epochs  \\
\midrule
VQA & 2e-5/5e-6 & 1024 &  8 \\
Captioning$\dagger$ & 1e-5\&8e-7 & 256& 5 \\
Retrieval & 1e-5/2e-6 & 256& 5 \\
Visual Grounding & 2e-5/2e-6 & 512& 120 \\
\bottomrule[1.5pt]
\end{tabular} 
\caption{Finetuning hyperparameters for downstream tasks. $\dagger$ denotes two stages fine-tuning.}
\label{table:finetune-hyper}
\end{table}

\paragraph{Image Captioning.} Image captioning requires generating a descriptive and fluent caption for a given image. We evaluate the performance of \modelname on two popular datasets: COCO Caption~\cite{Lin2014MicrosoftCC} and NoCaps~\cite{nocaps}. We fine-tune \modelname on the training set of COCO Caption and test it on the same Karpathy split~\cite{Li2020OscarOA,Wang2021SimVLMSV} as well as the NoCaps validation set. To fine-tune \modelname on COCO Caption, we follow the approach in~\cite{Li2020OscarOA} and first train the model with cross-entropy loss for 5 epochs with a learning rate of 1e-5 and a batch size of 256. We then further fine-tune the model with CIDEr optimization~\cite{scst} for an additional 5 epochs with a smaller learning rate of 8e-7. We use the best checkpoint on COCO Caption to predict on the NoCaps validation set.
During inference, we use beam search with a beam size of 10 and set the maximum generation length to 20.

% \paragraph{Image-Text Retrieval.} We conduct experiments for both image-to-text retrieval (TR) and text-to-image retrieval (IR) on COCO ~\cite{Lin2014MicrosoftCC} and Flickr30K ~\cite{Plummer2015Flickr30kEC} datasets. We adopt the widely-used Karpathy split ~\cite{karpathy2015deep} for both COCO and Flickr30K. COCO contains 113k/5k/5k images for train/validation/test, and Flickr30K contains 29k/1k/1k images for train/validation/test. Following ~\cite{Li2021AlignBF, li2022blip}, we jointly optimize the ITC loss and the ITM loss during fine-tuning. During inference, we first select top-k candidates by computing the dot-product similarity between the image and text encoder features (When extracting the image encoder feaure, for efficiency of coarse-grained ranking, we replace the SPD with a simple strategy which we directly detect the patch in a transformer layer based on the self-attention weights of image [CLS] token to other patch tokens ),and then rerank the selected candidates based on their ITM scores (In the fine-grained reranking stage, for the same image, we re-extracting multiple image encoder features based on SPD with the guidence of multiple text candidates.). We set $k = 256$ for COCO and $k = 128$ for Flickr30K.
\paragraph{Image-Text Retrieval.} We conducted experiments on both image-to-text retrieval (TR) and text-to-image retrieval (IR) using the COCO~\cite{Lin2014MicrosoftCC} and Flickr30K~\cite{Plummer2015Flickr30kEC} datasets and used the widely-used Karpathy split~\cite{karpathy2015deep} for both. COCO contains 113k/5k/5k images for train/validation/test, while Flickr30K contains 29k/1k/1k images for train/validation/test. During fine-tuning, we jointly optimized the ITC loss and the ITM loss following the approach in~\cite{Li2021AlignBF, li2022blip}. During inference, we first selected the top-k candidates by computing the dot-product similarity between the image and text encoder features (We set $k=256$ for COCO and $k=128$ for Flickr30K). For efficiency of coarse-grained ranking, we directly set $\beta$ to 0 and selected the patch based on the attention weights of the image [CLS] token to other patch tokens. During the fine-grained reranking for the top-k candidates, we set $\beta$ to 0.8 and reranked the candidates based on their ITM scores.

\paragraph{Visual Grounding.} The task of visual grounding involves localizing the referred object in an image given a plain text query. Instead of directly regressing bounding boxes, our approach concatenates visual features with textual features, which are then fed into the multi-modal decoder to predict the object's coordinates. We evaluate our method on the referring expression grounding dataset: RefCOCO+\cite{yu2016modeling}. The RefCOCO+ dataset contains 19K images and 141K queries.

\begin{figure*}[htpb]
     \centering
     \includegraphics[width=0.97\textwidth]{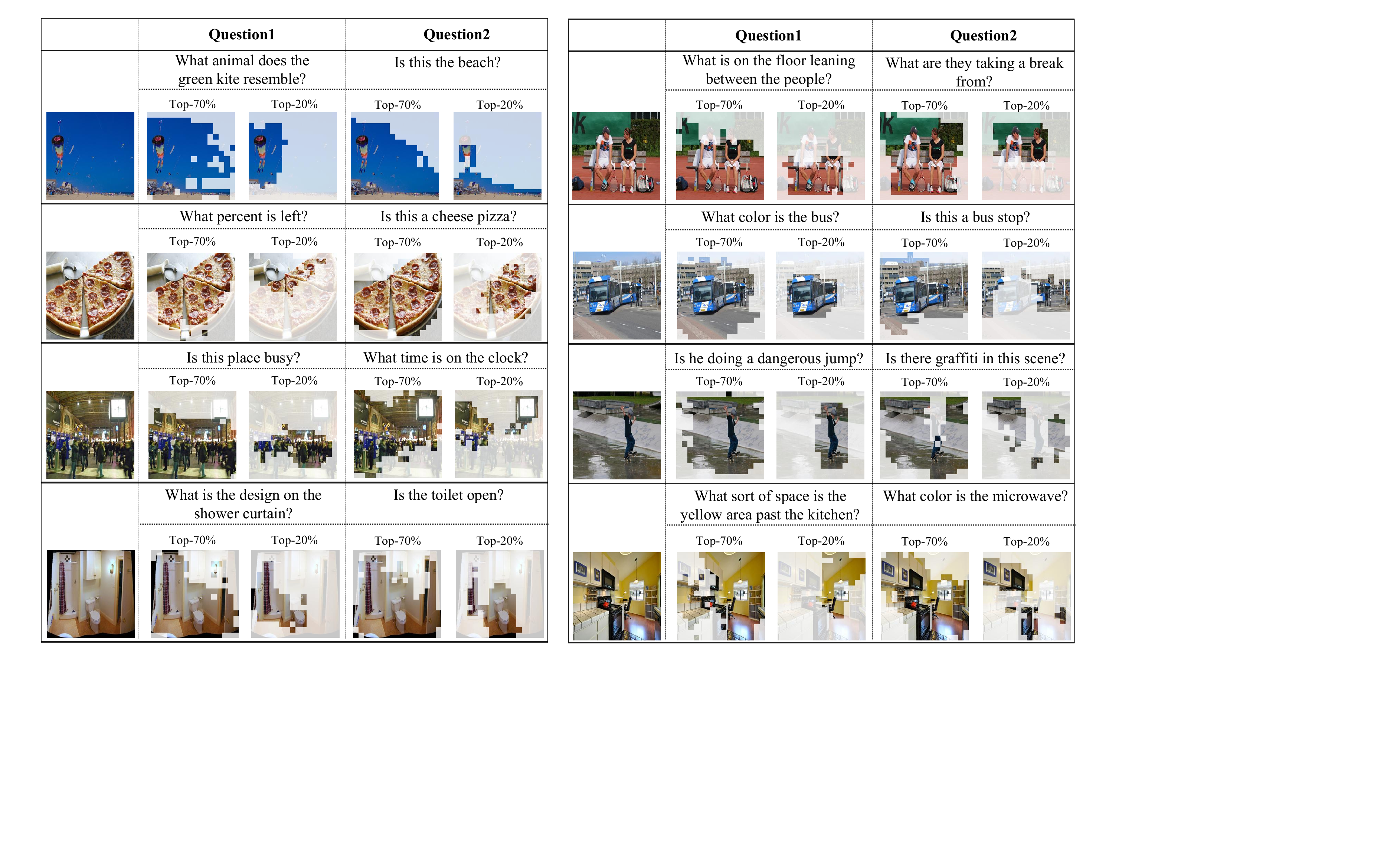}
     \caption{The visualization of the VQA cases and the selected text-relevant image patches.}
     \label{fig:vqa_case}
 \vspace{-1ex}
\end{figure*}
\section{Comparison Models}
\label{sup:comparison models}
\begin{itemize}

\item \textbf{E2E-VLP}~\cite{Xu2021E2EVLPEV}: proposes the first end-to-end VLP method for both V+L understanding and generation, with a unified Transformer encoder-decoder architecture.

\item \textbf{VinVL}~\cite{2021VinVL}: pre-trains a large-scale object-attribute detection model with much larger amounts of supervised data on four public object detection datasets for extracting better region-based visual feature. 

\item \textbf{OSCAR}~\cite{Li2020OscarOA}: proposes to use object tags detected in images as anchor points to ease the learning of cross-modal alignments, where the input to the Transformer is a combination of image, text and object tags.

\item \textbf{METER}~\cite{dou2021empirical}: systematically investigates how to design and pre-train a fully transformer-based VL model in an end-to-end manner.

\item \textbf{VLMo}~\cite{Wang2021VLMoUV}: presents a unified vision-language pretrained model that jointly learns a dual encoder and a fusion encoder with a modular Transformer network.

\item \textbf{SimVLM}~\cite{Wang2021SimVLMSV}: different from previous VLP methods that only use limited (4M-10M) image-text pairs for pre-training, it proposes a simple VLP model with a single prefix language modeling objective, which pre-trains on a extremely large aligned cross-modal data of about 1.8B noisy image-text pairs. This is also a latest state-of-the-art method on image captioning.

\item \textbf{ALBEF}~\cite{Li2021AlignBF}: introduces a contrastive loss to align the image and text representations before fusing them through cross-modal attention, which enables more grounded vision and language representation learning.

\item \textbf{UNITER}~\cite{Chen2020UNITERUI}: proposes an improved single-stream VLP method, by designing two new pre-training strategies: 1) it uses conditional masking on pre-training tasks instead of random masking strategy, 2) it designs a new word-region alignment pre-training task via the use of optimal transport to explicitly encourage fine-grained alignment between words and image regions. 

\item \textbf{ALIGN}~\cite{jia2021scaling}: leverages a noisy dataset of over one billion image alt-text pairs, obtained without expensive filtering or post-processing steps in the Conceptual Captions dataset.

\item \textbf{VLBERT}~\cite{Su2020VLBERTPO}: is a pioneering work to pre-train a single-stream multi-modal Transformer, which jointly trains both the Transformer-based cross-modal fusion and Fast R-CNN image feature extractor in both pre-training and fine-tuning phases. It is widely used as a baseline method for VLP models.

\noindent \textbf{VILT}~\cite{Kim2021ViLTVT}:  adopts linear projection and word embedding as the visual and textual encoders, and uses the visual transformer as the cross-modal encoder to align and fuse the features of both modalities in an end-to-end manner.

\item \textbf{VILLA}~\cite{gan2020large}: is the first known effort on large-scale adversarial training for vision-and-language (V+L) representation learning.

\item \textbf{XVLM}~\cite{Zeng2021xvlm}: proposes to learn multi-grained alignments which locates visual concepts in the image given the associated texts, and in the meantime align the texts with the visual concepts.
\item \textbf{BLIP}~\cite{li2022blip}: proposes a new VLP framework which transfers flexibly to both vision-language understanding and generation tasks. It effectively utilizes the noisy web data by bootstrapping the captions.
\item \textbf{UNICORN}~\cite{DBLP:journals/corr/abs-2111-12085}: proposes a vision-language (VL) model that unifies text generation and bounding box prediction into a single architecture. 

\item \textbf{LXMERT}~\cite{Tan2019LXMERTLC}: is the pioneering work to pre-train a two-stream multi-modal Transformer, which consists of an object relationship encoder, a language encoder and a cross-modality encoder. It is widely used as a baseline method for VLP models. 

\item \textbf{ViLBERT}~\cite{Lu2019ViLBERTPT}: proposes one of the first work that extend the BERT architecture to a multi-modal two-stream VLP model, which processes both visual and textual inputs in separate streams that interact through co-attentional transformer layers.

\item \textbf{mPLUG} ~\cite{li2022mplug}: is a  vision-language foundation model for both cross-modal understanding and generation and introduces an effective and efficient vision-language architecture with novel cross-modal skip-connections.
\item \textbf{TRIPS} ~\cite{Jiang2022TRIPSEV}: is a vision-and-language pre-training model which reduces the visual sequence progressively with a patch-selection layer in the visual backbone for efficient training and inference.
\end{itemize}
\section{Case Study }
\label{sup:case study}
In this subsection, we visualize more VQA cases and the selected text-relevant image patches in Figure~\ref{fig:vqa_case}. Note that these two examples are not cherry-picked. The phenomenon in these examples is commonly observed among other samples. 
% {\small
% \bibliographystyle{ieee_fullname}
% \bibliography{egbib}
% }

\end{document}